\documentclass[10pt,journal,compsoc]{IEEEtran}
\ifCLASSOPTIONcompsoc
\usepackage[nocompress]{cite}
\usepackage{amsmath,amsfonts}
\usepackage{algorithm}
\usepackage{algpseudocode}
\usepackage{array}
\usepackage[caption=false,font=normalsize,labelfont=sf,textfont=sf]{subfig}
\usepackage{textcomp}
\usepackage{stfloats}
\usepackage{url}
\usepackage{verbatim}
\usepackage{graphicx}
\usepackage{cite}
\usepackage{cases}
\usepackage{bm}
\usepackage{amssymb}
\usepackage{multirow}
\usepackage{epstopdf}
\usepackage{epsfig}
\usepackage{caption}
\usepackage{makecell}
\usepackage{longtable}
\usepackage{booktabs}
\usepackage{tabularx}
 \usepackage{amsthm}
 \usepackage{color}
 \usepackage{xcolor}
\usepackage{adjustbox}

\newtheorem{thm}{\bf Theorem}

\newtheorem{rmk}{\bf Remark}

\newenvironment{remark}{\begin{rmk}} {\end{rmk}}
\else
  \usepackage{cite}
\fi
\ifCLASSINFOpdf
\else
\fi
\usepackage{amsmath,graphicx}

\begin{document}
%
\title{Curvature regularization for Non-line-of-sight Imaging from Under-sampled Data}
%
%
%
%

\author{Rui Ding, Juntian Ye, Qifeng Gao, Feihu Xu,
        and Yuping Duan

\thanks{The work was supported by the National Natural Science Foundation of China NSFC 12071345. }
\IEEEcompsocitemizethanks{ 
\IEEEcompsocthanksitem R. Ding and Q. Gao are with the Center for Applied Mathematics, Tianjin University, Tianjin 300072, China. E-mail: \{rding,gaoqifeng$\_$98\}@tju.edu.cn
\IEEEcompsocthanksitem J. Ye and F. Xu are with Hefei National Laboratory for Physical Sciences at Microscale and School of Physical Science, University of Science and Technology of China, Hefei 230026, China. E-mail: jt141884@mail.ustc.edu.cn;~feihuxu@ustc.edu.cn
\IEEEcompsocthanksitem Y. Duan is with the School of Mathematical Sciences, Beijing Normal University, Beijing, 100875, China.
E-mail: doveduan@gmail.com}
\thanks{Manuscript received April 19, 2005; revised August 26, 2015.}}

%
%

\markboth{Journal of \LaTeX\ Class Files,~Vol.~14, No.~8, August~2015}%
{Shell \MakeLowercase{\textit{et al.}}: Bare Demo of IEEEtran.cls for Computer Society Journals}
%



\IEEEtitleabstractindextext{%
\begin{abstract}
Non-line-of-sight (NLOS) imaging aims to reconstruct the three-dimensional hidden scenes from the data measured in the line-of-sight, which uses photon time-of-flight information encoded in light after multiple diffuse reflections. The under-sampled scanning data can facilitate fast imaging. However, the resulting reconstruction problem becomes a serious ill-posed inverse problem, the solution of which is highly possibility to be degraded due to noises and distortions. In this paper, we propose novel NLOS reconstruction models based on curvature regularization, 
i.e., the object-domain curvature regularization model and the dual (signal and object)-domain curvature regularization model. In what follows, we develop efficient optimization algorithms relying on the alternating direction method of multipliers (ADMM) with the backtracking stepsize rule, for which all solvers can be implemented on GPUs. We evaluate the proposed algorithms on both synthetic and real datasets, which achieve state-of-the-art performance, especially in the compressed sensing setting. Based on GPU computing, our algorithm is the most effective among iterative methods, balancing reconstruction quality and computational time. All our codes and data are available at https://github.com/Duanlab123/CurvNLOS. 
\end{abstract}

\begin{IEEEkeywords}
Non-line-of-sight, under-sampled scanning, curvature regularization, dual-domain reconstruction, GPU implementation
\end{IEEEkeywords}}

\maketitle

\IEEEdisplaynontitleabstractindextext

%
\IEEEpeerreviewmaketitle

\IEEEraisesectionheading{\section{Introduction}\label{sec:introduction}}

\IEEEPARstart{N}{on-line-of-sight} (NLOS) imaging uses time-resolved measurements to recover the 3D shape and visual appearance of hidden objects beyond the direct line of sight of sensors \cite{faccio2020non,geng2021recent}, which has various applications such as autonomous driving \cite{scheiner2020seeing}, 3D human pose estimation \cite{isogawa2020optical}, sensor system \cite{chen2020learned,zhu2021single}, and many other domains. 
Recent advances in single-photon detector technology and computational algorithms for solving large-scale inverse problems make NLOS imaging feasible in different conditions. For instance, NLOS imaging and real-time tracking of hidden objects have been demonstrated over a distance of 1.43 km \cite{wu2021non} and at a resolution of ~0.6 mm at a distance of 0.55 m \cite{cao2022high,PhysRevLett.127.053602}, respectively. 

The inverse problem of reconstructing the 3D shape and appearance of the hidden object is also very challenging \cite{o2018confocal,lindell2020three}. Roughly speaking the NLOS imaging reconstruction methods can be divided into three categories, i.e., direct reconstruction methods, iterative reconstruction methods, and deep learning-based reconstruction methods. The filtered back-projection algorithm is a kind of fast direct method by filtering the data and performing the back-projection operation, where the data is painted back in the image along the direction it is being measured \cite{gupta2012reconstruction,la2018error,arellano2017fast}. Other direct reconstruction methods also contain the phasor field \cite{liu2019non}, frequency-domain method \cite{lindell2019wave}, Fermat flow method \cite{xin2019theory}, etc, which are proposed by modeling the physical process of imaging. These direct reconstruction methods are fast and efficient but are sensitive to noises and measurement distortions.  
The iterative reconstruction methods have been used by introducing priors to regularize the NLOS reconstruction problem and improve the image quality \cite{HeideXHH14,AhnDVGS19,liu2021non,ye2021compressed,LiuZHLX22}. Although iterative reconstruction methods can provide high-quality reconstruction images, they also consume much more computational time than direct methods. Due to the development of deep learning, convolutional neural network models have been developed for solving the NLOS reconstruction problem \cite{ChopiteH0I20,ChenWKRH20,Isogawa0OK20,shen2021non,zhu2023compressive}. However, deep learning methods are highly dependent on the training datasets, which makes it possible to lose their impact on real measurement data and spatial/temporal degradation data \cite{ChopiteH0I20}. 

Indeed, the dense raster scanning used in the aforementioned methods is detrimental to high-speed NLOS applications. Thus, different strategies have been employed to reduce the acquisition time. One strategy is to use the multi-pixel time-of-flight NLOS system. 
Nam \emph{et al.} \cite{nam2021low} used the specifically designed single photon avalanche diode (SPAD) array detectors to realize a multi-pixel NLOS imaging method together with a fast reconstruction algorithm that can capture and reconstruct live low-latency
videos of NLOS scenes.  
Pei \emph{et al.} \cite{pei2021dynamic} used the SPAD array and an optimization-based computational method to achieve NLOS reconstruction of 20 frames per second.  Another strategy is to use fewer scanning points to reconstruct the scene. Isogawa \emph{et al.} \cite{isogawa2020efficient} proposed a circular and confocal non-line-of-sight scan, for which the scanning involves sampling points forming a circle on a visible wall to reduce the dimension of transient measurements. Ye \emph{et al.} \cite{ye2021compressed} explored compressed sensing in active NLOS imaging to reduce the required number of scanning points for fast implementation. Zhang \emph{et al.} \cite{zhang2023non} proposed an Archimedean spiral scanning method based on confocal non-line-of-sight imaging, which greatly reduces the data acquisition time. 

As illustrated in previous works \cite{isogawa2020efficient,ye2021compressed}, NLOS reconstruction can be realized by under-sampled measurements to facilitate high-speed acquisition. However, sparse measurements may lead to the degradation of reconstructed images. Effective regularization methods have been used for various shape and image processing tasks, such as sparse regularization \cite{liu2022rank}, surface area regularization \cite{liu2019surface}, and curvature regularization \cite{ChambolleP19,DongJLS20,ZhongYD21}, etc. Among them, curvature regularization is well-known for its ability to model the continuities of edges and surfaces. 
Initially, curvature regularization was applied to the problem of image inpainting to restore satisfactory results meeting human perception  \cite{MasnouM98,zhong2020minimizing,SchraderAWE22}. Due to its superiority in recovering missing data, curvature regularity has also been used for surface reconstruction \cite{LeflochKSWK17,HeKL20}, sparse image reconstruction \cite{zheng2018few,YanD20}.  
Since the curvature regularization is capable of capturing the tiny but elongated structures in images, it has also been used for image segmentation \cite{ZhuTC13}, tubular structure tracking \cite{LiuCSLSPC22}, etc. Curvature regularization is a good choice for under-sampled NLOS imaging problems to obtain smooth and satisfied reconstructed surfaces.

In this paper, we study the curvature regularization reconstruction model for under-sampled NLOS imaging problems, called curvNLOS, which can reconstruct surfaces with good quality through as few measurements as possible. We also develop an effective numerical algorithm based on the alternating direction methods of multipliers (ADMM), which can be solved by GPU computation. Comprehensive experiments are conducted on both synthetic and real data. The numerical results demonstrate that curvature regularization can effectively guarantee the smoothness of the object and estimated signals. The main contribution of this work can be summarized as follows:
\begin{itemize}
    \item We propose novel curvature regularization models for solving the NLOS imaging problem, where the curvature regularization is used to restore the smooth surface of hidden objects and fill in the sparse measured signals.
    \item We present fast iterative optimization algorithms for solving the curvature minimization problems, where the high-order curvature is regarded as the adaptive weight for total variation, and the linearization technique is used to obtain the GPU-friendly iterative algorithms.
    \item We develop a GPU-based NLOS reconstruction package by utilizing the parallel computation ability of GPUs, which is desirable for high-speed NLOS applications. 
    \item By comparing with the state-of-the-art NLOS methods, our curvature regularization models can robustly restore estimated signals and the three-dimensional scene points on both confocal and non-confocal settings, especially when the measurements are under-sampled measured data.
\end{itemize}


\section{Forward propagation model}
\label{sec2}
In transient imaging, a time-resolved detector is used to measure the incident flux of photons as a function of emitted light impulses. Let $\bm  x= (x,y,z)$ be the three-dimensional scene coordinates, and $\bm x'_i= (x'_i,y'_i,z=0)$, $\bm x'_d= (x'_d,y'_d,z=0)$ be the illumination and detection coordinates on the visual wall, respectively. 
The light emitted by the laser passes through the scanning galvanometer and hits the illumination point $x'_i$, and diffuses towards the target point $\bm x$. Then, the photons reflect on the target and propagate to the detection point $\bm x'_d$. Finally, the photons diffuse back and enter the single-pixel detectors such as SPAD. The NLOS reconstruction aims to recover the location, shape, albedo, and normal of the target from the detected number of photons. 

The general non-confocal direction-albedo forward propagation model for NLOS imaging can be described below
\begin{equation*}
\label{directional_nonconfocal_model}
\begin{split}
\tau(\bm x^\prime_i,\bm x^\prime_d, t)= &\iiint_{\Omega} \frac{(\bm x^\prime_i-\bm x)\cdot \bm n(\bm x)}{d(\bm x^\prime_i,\bm x)^3}\cdot\frac{(\bm x^\prime_d-\bm x)\cdot \bm n(\bm x)}{d(\bm x^\prime_d, \bm x)^3}u(\bm x)\\
&\cdot\delta \Big(d(\bm x^\prime_i,\bm x)+d(\bm x^\prime_d,\bm x)-tc\Big)d\bm x,
\end{split}
\end{equation*}
where $\tau$ is the recorded transient image, $u$ is the albedo of the hidden scene at each point $\bm x$ with $z>0$ in the 3D half-space $\Omega$, $\bm n(\bm x) = (n_x,n_y, n_z)(\bm x)$ denotes the surface normal. The $\delta$ represents the surface of a spatio-temporal four-dimensional hypercone defined by the equation $x^2 + y^2 + z^2 - (tc/2)^2 = 0$. It models light propagation from the wall to the object and back to the wall, converting time $t$ to the travel distance $tc$, where $c$ is the speed of light. Studies on non-confocal NLOS imaging can be found in \cite{tsai2017geometry,tsai2019beyond}. 
The distances between the reconstructed point and the illumination point and detection point are defined as   
\[d(\bm x^\prime_i,\bm x)=\sqrt{(x^\prime_i-x)^2+(y^\prime_i-y)^2+z^2},\]
and 
\[d(\bm x^\prime_d,\bm x)=\sqrt{(x^\prime_d-x)^2+(y^\prime_d-y)^2+z^2},\]
respectively. 
When the illumination point and detection point are located at the same position, i.e., $\bm x^\prime= \bm x^\prime_i=\bm x^\prime_d$,  we have the confocal direction and albedo NLOS reconstruction model as follows
\begin{equation}
\label{directional_confocal_model}
\tau(\bm x^\prime, t)= \iiint_{\Omega} \frac{u(\bm x)\bm n(\bm x)}{d(\bm x^\prime, \bm x)^4}\cdot \bm n(\bm x)\cdot\delta \Big(2d(\bm x^\prime, \bm x)-tc\Big)d\bm x.
\end{equation}
Methods to localize the three-dimensional scene points and estimate their surface normals of hidden objects have been established for the NLOS imaging \cite{heide2019non,young2020non,liu2021non}. 
As long as $\langle\bm n(\bm x), \bm n(\bm x) \rangle = 1$, the direction-albedo forward model reduces into the volumetric albedo model 
\begin{equation}
\label{albedo_confocal_model}
\tau(\bm x^\prime, t)= \iiint_{\Omega} \frac{u(\bm x)}{d(\bm x^\prime, \bm x)^4}\cdot\delta \Big(2d(\bm x^\prime, \bm x)-tc\Big)d\bm x.
\end{equation}
The corresponding discrete image formation of the NLOS model \eqref{albedo_confocal_model} is given as 
\begin{equation}\label{linear_prob}
    \tau = Au = R_t^{-1} HR_z u,
\end{equation}
where the matrix $A=R_t^{-1}H R_z$ is the light transport matrix, $H$ represents the shift-invariant 3D convolution, $R_t$ and $R_z$ represent the transformation operations applied to the temporal and spatial dimensions, respectively.

The task of recovering the object $u$ from the measured signal $\tau$ is a typical ill-posed inverse problem. The regularization method is a good choice for solving ill-posed inverse problems. In \cite{ye2021compressed}, the sparsity regularization and a non-negative constraint were used to restore the surface of objects. Although the algorithm can reconstruct a three-dimensional hidden image of $64\times64$ spatial resolution with $5\times5$ scanning points, the results greatly depend on the post-processing step to smooth out the noises and outliers. The collaborative signal and objective regularization was used in \cite{liu2021non}, which is shown effective on both confocal and non-confocal NLOS imaging reconstruction. However, such complex regularization makes the computational costs increase extremely, which is unfavorable for real applications.  Thus, finding suitable regularization methods for NLOS imaging algorithms can consider both imaging quality and speed, which is still a challenging problem.

\begin{figure}
	\centering
        \captionsetup[subfloat]{labelsep=none,format=plain,labelformat=empty}
        
    ~~\subfloat[\footnotesize{(a)Ground truth}]{\includegraphics[width=1.0in,height=25mm]{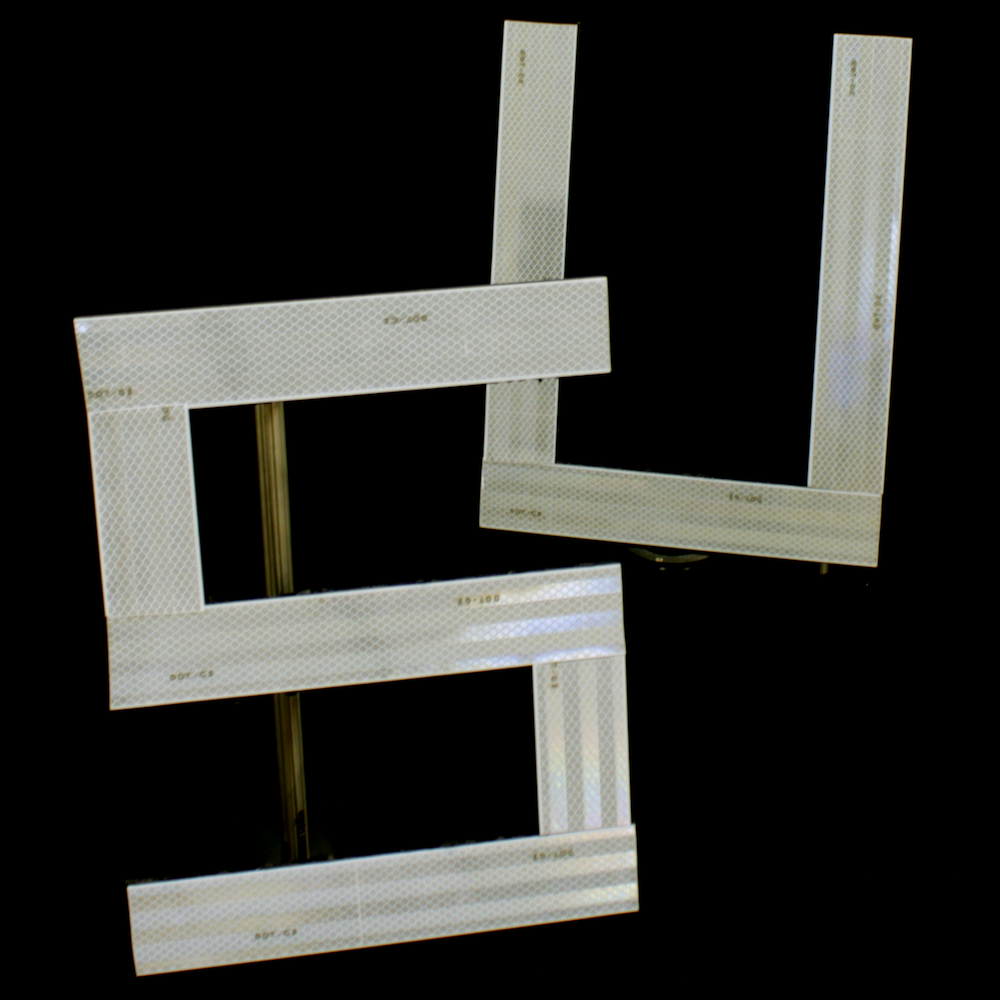}}~
    \subfloat[\footnotesize{(b)$\tau (32,32,:)$}]{\includegraphics[width=1.0in,height=25mm]{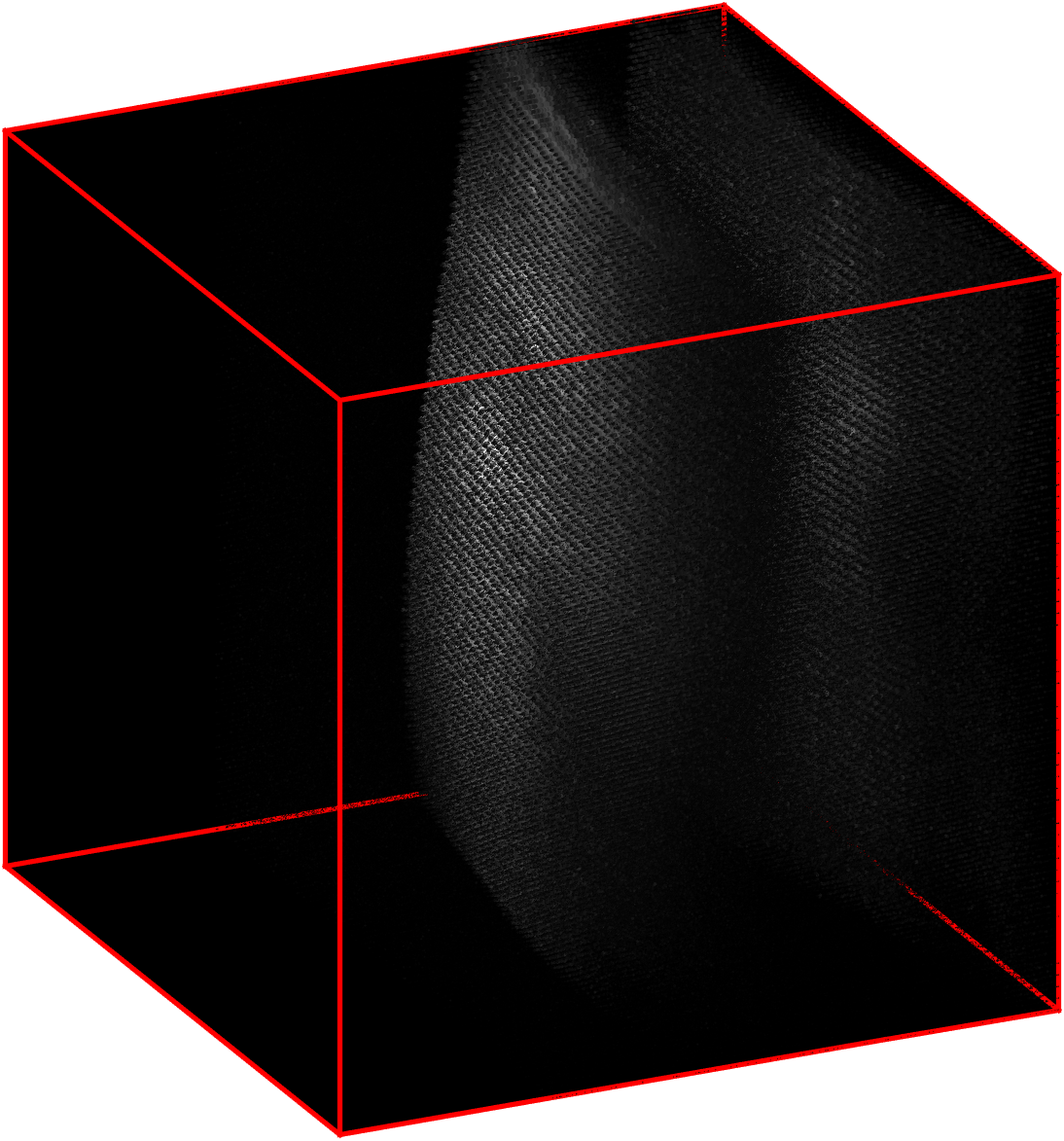}}~
    \subfloat[\footnotesize{(c)$\tau (32,32,:)$}]{\includegraphics[width=1.2in,height=25mm]{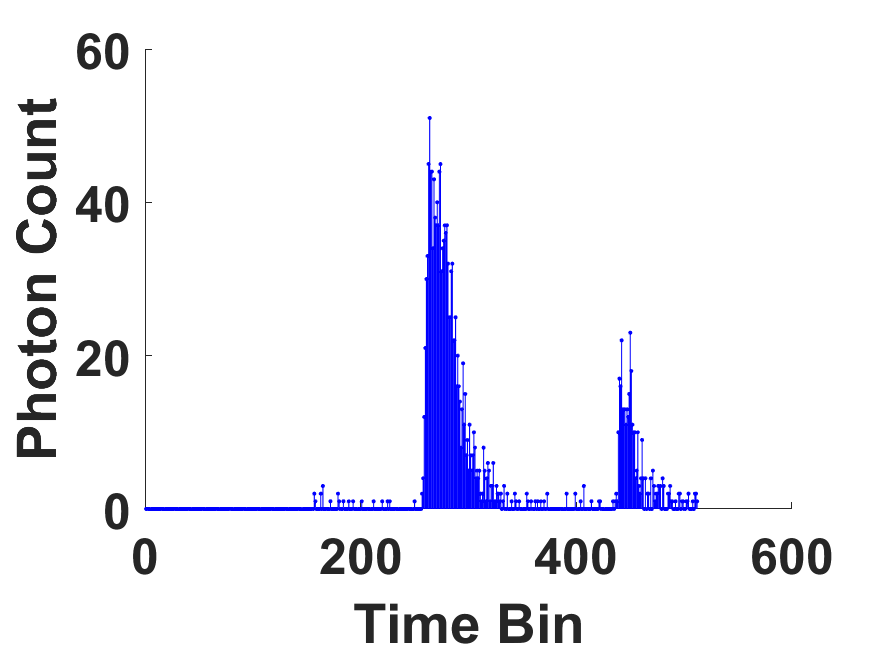}}\\
    \vspace{0.02cm}
    \caption{Overview of NLOS imaging measurements, where (a) The ground truth of hidden objects, (b) The measurements of the wall $\tau$ attenuated along the time axis, and (c) A histogram measured at a selected scanned point on the visible wall.}
	\label{fig2}
\end{figure}

\section{Object-domain curvature method}
\label{sec3}
As shown in Fig. \ref{fig2} (b), the 3D transient measurement $\tau(x,y,t)$ measured by a time-resolved detector is a time-continuous function, in which the number of photons or intensity is recorded at each arrival time. Fig. \ref{fig2} (c) is a histogram measured at a selected scanning point on the visible wall indicating the temporal precision of the detector. The curvature regularization term can provide strong prior information on the continuity of the image to naturally fill in the missing information and obtain a smooth surface. Thus, we use the curvature regularization to approximate the oracle signal corresponding to the real hidden scene. 
To be specific, we aim to minimize the following curvature-related energy for the NLOS imaging problem 
\begin{equation}\label{image_model}
\min\limits_{u} \Big\{E(u) =
\frac12\|Au-\tau_0\|^2_{\Omega \backslash X} + \mathcal R(\kappa(u))\Big\},
\end{equation}
where $X\subset \Omega$ denotes the missing region, and $\mathcal R(\cdot)$ denotes the curvature regularization. More specifically, we define the regularization term as follows
\[\mathcal R(\kappa(u)) = \sum_{\bm x}\phi(\kappa(u(\bm x))) |\nabla u(\bm x)|,\]
where $|\nabla u|$ is the total variation of $u$ and $\kappa$ denotes the curvature defined by 
\[\kappa = \nabla\cdot\Big(\frac{\nabla u}{|\nabla u|}\Big),\]
and $\phi(\cdot)$ is the function of $\kappa$ defined similar to \cite{zhong2020minimizing} as
\begin{equation*}
\phi(\kappa)= \begin{cases}a+b|\kappa|, & \text {Total Absolute Curvature (TAC),} \\ 
\sqrt{a+b|\kappa|^2}, & \text {Total Roto-translation Variation (TRV),} \\ a+b|\kappa|^2, & \text {Total Squared Curvature (TSC),}\end{cases}
\end{equation*}
with $a$ and $b$ being two positive constants. Here, when $b=0$, it gives us the classical total variation (TV) regularization.

\subsection{Numerical minimization}
To efficiently solve the curvature regularization model \eqref{image_model}, we regard curvature terms as the weights of the total variation. Therefore, we reformulate it into a constrained minimization problem as follows 
\begin{equation*}
\begin{split}
  \min\limits_{u,v}~&~\frac{1}{2}\|Au-\tau_0\|^2_{\Omega \backslash X} + \sum_{\bm x}\phi(\kappa(u(\bm x))) |v(\bm x )|,  \\
  \mbox{s.t.}~& ~~ v =\nabla u.
  \end{split}
\end{equation*}
Then the associated augmented Lagrangian functional can be defined as follows
\begin{equation}
\begin{split}
\mathcal L (u,v;\Lambda) = \frac{1}{2}\|Au&-\tau_0\|^2_{\Omega \backslash X} +\sum_{\bm x}\phi(\kappa(u(\bm x))) |v(\bm x)| \\&+ \langle \Lambda,v-\nabla u\rangle + \frac{\mu}{2} \|v-\nabla u \|^2_2,
\end{split}
\label{spp1}
\end{equation}
where $\Lambda$ is the Lagrange multiplier, and $\mu$ is the positive parameter. 
We use ADMM to iterative and alternatively solve the sub-minimization problems.  Now we discuss the sub-minimization problems for solving \eqref{spp1} and the corresponding solutions.  

\subsubsection{Sub-minimization problem w.r.t. $v$}
The sub-minimization problem w.r.t. $v$ is defined as
\begin{equation}
\min\limits_{v}~ \sum_{\bm x}\phi(\kappa(u^{k}(\bm x))) |v(\bm x)|+ \frac{\mu}{2} \big\|v - (\nabla u^{k} -\frac{\Lambda^k}{\mu})\big\|^2_2.    
\label{vsub}
\end{equation} 
Since $\phi(\kappa(u^k(\bm x)))$ is known in advance, the above minimization problem becomes the weighted $\ell_1$ and $\ell_2$ minimization problem, which can be solved by the shrinkage operator as follows 
\begin{equation}\label{v-sol}
v^{k+1} = \mbox{shrinkage}\bigg(\nabla u^{k} - \frac{\Lambda^k}{\mu},\frac{\phi(\kappa(u))}{\mu}\bigg),
\end{equation}
with the shrinkage operator defined by 
$\mbox{shrinkage}(a,b) = \max\{|a|-b,0\}\circ \frac{a}{|a|}$.

\subsubsection{Sub-minimization problem w.r.t. $u$}
The $u$ sub-minimization problem can be formulated as a quadratic minimization problem 
\begin{equation}
\mathop{\min}\limits_{u}~\frac{1}{2}\big\|A_D u-\tau_0 \big\|^2_2 + \frac{\mu}{2} \big\|\nabla u -(v^{k+1}+\frac{\Lambda^k}{\mu})\big\|^2_2,
\label{usub}
\end{equation}
where $A_D = DA$ with $D$ being the sampling operator to define the missing data domain. For simplicity, we denote $f(u)=\frac{1}{2}\big\|A_D u-\tau_0 \big\|^2_2, ~g(u)=\frac{\mu}{2} \big\|\nabla u -(v^{k+1}+\frac{\Lambda^k}{\mu})\big\|^2_2.$ 
Then we reformulate the minimization problem \eqref{usub} using a quadratic approximation of $f(u)$ at a given point $u_k$ as follows
\begin{equation}
 \mathop{\min}\limits_{u}~f(u_k) + \langle u-u_k,\nabla f(u_k)\rangle + \frac{L}{2}\big\|u-u_k \big\|^2_2 + g(u),
\end{equation}
which is equivalent to 
\begin{equation}
 \mathop{\min}\limits_{u}~ g(u) + \frac{L}{2}\big\|u-(u_k - \frac{1}{L} \nabla f(u_k)\big\|^2_2,
\label{proximal}
\end{equation}
with $L$ being $2\|A_D\|^2_2$. 
The optimal value of the \eqref{proximal} becomes the solution to the following partial differential equation
\begin{equation*}
\label{tau-pde}
\begin{split}
(L\mathcal I + \mu\nabla^*\nabla )u = Lu_k - \nabla f(u_k) + \mu\nabla^*(v^{k+1}+\frac{\Lambda^k}{\mu}),
\end{split}
\end{equation*}
where $\nabla f(u_k) = A_D^T(A_Du_k-\tau_0)$ and $\nabla^*$ is the adjoint operator of gradient. The above PDE can be solved by the Fast Fourier Transform (FFT) as follows
\begin{equation}\label{u-sol}
u^{k+1} = \mathcal F^{-1}\bigg(\frac{\mathcal F\big(Lu_k - \nabla f(u_k) + \mu\nabla^*(v^{k+1}+\frac{\Lambda^k}{\mu})\big)}{L\mathcal I + \mu \mathcal F (\nabla^*\nabla)}\bigg),
\end{equation}
where $\mathcal F$ and $\mathcal F^{-1}$ denote the forward and inverse FFT operation, respectively.

\subsubsection{Update of Lagrange multipliers}
Finally, we update the Lagrange multipliers by the gradient ascend method as follows
\begin{equation}
\Lambda^{k+1} = \Lambda^{k} + \mu (v^{k+1} - \nabla u^{k+1}).
\end{equation}

\subsection{Our algorithm}
We summarize the ADMM-based algorithm for solving the object-domain curvature regularization model as Algorithm \ref{alg1}. As seen, the acceleration technique in \cite{beck2009fast} is applied to obtain a faster convergence rate, where the specific linear combination of the previous two points $\{u^{k-1},u^{k}\}$ is used in the computation. 

\begin{algorithm}
\caption{\label{alg1}The ADMM-based algorithm for solving  \eqref{image_model}}
\begin{algorithmic}[h]
\State \textbf{Input}: Raw data $\tau_0$ and parameters $\Lambda$, $a$, $b$, $\mu$, $\epsilon$, $\bar u^0 =u^0=0$, $\Lambda^0 = 0$, $T_{max}$, $t^0=1$;
\For{k=1} {$T_{max}$}
    \Statex /* Solve the saddle-point problem */
    \State $(u^{k+1},v^{k+1};\Lambda^{k+1}) = \max\limits_{\Lambda}\min\limits_{u,v} \mathcal{L}(\bar u^{k},v^k;\Lambda^k)$;
    \Statex /* Compute $t^{k+1}$ and $\bar u^{k+1}$ */
    \State $t^{k+1} = \frac{1+\sqrt{1+4(t^k)^2}}{2}$;
    \State $\bar u^{k+1} = u^{k+1} + \frac{t^k-1}{t^{k+1}}(u^{k+1}-u^{k})$;
    \Statex /* Update $\phi(\kappa_u^{k+1})$ */
    \State $\phi(\kappa(u^{k+1})) = a + b(\nabla \cdot \frac{\nabla u^{k+1}}{|\nabla u^{k+1}|})^2$;
    \Statex /* Stopping condition */
    \If{$e^{k+1}\!=\!\big\|E(u^k)\!-\!E(u^{k+1})\big\|\big/\big\|E(u^{k+1})\big\| \leq \epsilon$}
    \State \textbf{Return} $u^{k+1}$;
    \EndIf
\EndFor
\State \textbf{Output}: $u^{k+1}$.
\end{algorithmic}
\end{algorithm}

\begin{remark}
Due to the non-convexity of the curvature regularization in our model \eqref{image_model}, the convergence of Algorithm \ref{alg1} is difficult to obtain theoretically. A partial convergence result can be found in our previous work \cite{zhong2020minimizing}. From observing the numerical energy decay, Algorithm \ref{alg1} numerically converges quite stable; see Fig. \ref{curve}.
\end{remark}

\section{Dual-domain curvature method and GPU implementation}
\label{sec4}
\subsection{Dual-domain reconstruction method}
Since the under-sampled signal can be regarded as the data inpainting problem, we also propose a dual-domain reconstruction model, where the curvature regularization is employed for both the signal domain and object domain. Mathematically, we present the dual-domain NLOS imaging reconstruction model as follows
\begin{equation}\label{dual_model}
\begin{split}
\min\limits_{u,\tau} \Big\{E(u,\tau) =\frac{1}{2}\|Au-\tau\|^2_2 &+ \frac{\lambda}{2}\|\tau-\tau_{0}\|^2_{\Omega \backslash X}\\ &+ \mathcal R(\kappa(u)) + \mathcal R(\kappa(\tau))\Big\},
\end{split}
\end{equation}
where $\lambda$ is the positive parameter and $\mathcal R(\kappa(\tau))$ is defined similar to $\mathcal R(\kappa(u))$ as 
\[\mathcal R(\kappa(\tau)) = \sum_{\bm x}\phi(\kappa(\tau(\bm x))) |\nabla \tau(\bm x)|.\]
Note that $\kappa(u)$ calculates the spatial curvature of the hidden object, while $\kappa(\tau)$ is used to measure the curvature of the time-dependent signal.
Similarly, we rewrite the above model into the following constrained minimization problem
\begin{equation*}
\begin{split}
  \min\limits_{u,\tau,v,w,f}&~ \sum_{\bm x}\phi(\kappa(u(\bm x))) |v(\bm x )| + \sum_{\bm x}\phi(\kappa(\tau(\bm x))) |w(\bm x )| \\&
  \qquad\qquad\qquad+ \frac{1}{2}\|Au-\tau\|^2_2 + \frac{\lambda}{2}\|f-\tau_{0}\|^2_{\Omega \backslash X}\\
  \mbox{s.t.}~~& ~~ v =\nabla u, ~~ w = \nabla \tau,~~ f = \tau.
  \end{split}
\end{equation*}
Then the associated augmented Lagrangian functional can be defined as follows
\begin{equation*}
\begin{split}
\mathcal L (u,\tau,v,&w,f;\Lambda_1,\Lambda_2,\Lambda_3) = \frac{1}{2}\|Au-\tau\|^2_2 + \frac{\lambda}{2}\|f-\tau_{0}\|^2_{\Omega \backslash X}\\&+\sum_{\bm x}\phi(\kappa(u(\bm x))) |v(\bm x)|+ \sum_{\bm x}\phi(\kappa(\tau(\bm x))) |w(\bm x)| \\&+ \langle \Lambda_1,v-\nabla u\rangle + \frac{\mu_1}{2} \|v-\nabla u \|^2_2 + \langle \Lambda_2,w-\nabla \tau \rangle \\&+ \frac{\mu_2}{2} \|w-\nabla \tau \|^2_2 + \langle \Lambda_3,f- \tau \rangle + \frac{\mu_3}{2} \|f- \tau \|^2_2,
\end{split}
\label{spp}
\end{equation*}
where $\Lambda_1$, $\Lambda_2$, $\Lambda_3$ are the Lagrange multipliers, and $\mu_1$, $\mu_2$, $\mu_3$ are the positive parameters. Then the ADMM can be implemented to iterative and alternatively solve the sub-minimization problems; see Algorithm \ref{alg2}. 

\begin{algorithm}
\caption{\label{alg2}The ADMM-based algorithm for solving \eqref{dual_model}}
\begin{algorithmic}[h]
\State \textbf{Input}: Raw data $\tau_0$, and parameters $a_u$, $b_u$, $a_\tau$, $b_\tau$, $\Lambda_1$, $\Lambda_2$, $\Lambda_3$, $\mu_1$, $\mu_2$, $\mu_3$, $\epsilon$, $\bar u^0 = u^0$, $\bar\tau=\tau=0$, $\Lambda_1^0 =\Lambda_2^0= \Lambda_3^0= 0$, $T_{max}$, $t^0=1$;
\For{k=1} {$T_{max}$}
\Statex /* Solve the saddle-point problem */
    \State $(u^{k+1},\tau^{k+1},v^{k+1},w^{k+1},f^{k+1};\Lambda_1^{k+1},\Lambda_2^{k+1},\Lambda_3^{k+1}) = \max\limits_{\Lambda_1,\Lambda_2,\Lambda_3}\min\limits_{u,\tau,v,w,f} \mathcal{L}(\bar u^{k},\bar\tau^{k},v^{k},w^{k},f^{k};\Lambda_1^{k},\Lambda_2^{k},\Lambda_3^{k})$;
    \Statex /* Compute $t^{k+1}$ and $\bar u^{k+1}$ */
    \State $t^{k+1} = (1+\sqrt{1+4(t^k)^2})/2;$
    \State $\bar u^{k+1} = u^{k+1} + \frac{t^k-1}{t^{k+1}}(u^{k+1}-u^{k});$
    \State $\bar \tau^{k+1} = \tau ^{k+1} + \frac{t^k-1}{t^{k+1}}(\tau ^{k+1}-\tau ^{k});$
    \Statex /* Update $\phi(\kappa(u^{k+1}))$ and $\phi(\kappa(\tau^{k+1}))$ */
    \State $\phi(\kappa(u^{k+1})) = a_u + b_u(\nabla \cdot \frac{\nabla u^{k+1}}{|\nabla u^{k+1}|})^2;$
    \State $ \phi(\kappa(\tau^{k+1})) = a_{\tau} + b_{\tau}(\nabla \cdot \frac{\nabla \tau^{k+1}}{|\nabla \tau^{k+1}|})^2;$
    \Statex /* Stopping condition */
    \If{$e^{k+1}=\big\|E(u^k)-E(u^{k+1})\big\|/\big\|E(u^{k+1})\big\| \leq \epsilon$}
    \State \textbf{Return} $u^{k+1}$;
    \EndIf
\EndFor
\State \textbf{Output}: $u^{k+1}$.
\end{algorithmic}
\end{algorithm}

The algorithm and solutions to the sub-minimization problems can be generalized from the object-domain cases. Both the object-domain reconstruction algorithm and dual-domain reconstruction algorithm are provided at https://github.com/Duanlab123/CurvNLOS. More details can be found in our public codes.

\begin{remark}
Note that we initialize the variable $u$ by Algorithm \ref{alg1} to obtain better convergence and high-quality reconstructions.
\end{remark}

\subsection{GPU implementation}
The GPU has a distinct advantage in parallel computing, consisting of thousands of smaller, more efficient cores designed for multitasking. The GPU-based image reconstruction allows for the use of more complex models and maintains reasonable execution time. Thus, we implement both Algorithm \ref{alg1} and Algorithm \ref{alg2} on the GPU to reduce the computational time. We utilized one TITAN RTX graphics card to run our algorithms, for which each iteration takes about 0.1 seconds for $128\times 128\times 512$ data. For data with higher dimensions, the advantage of GPU over CPU is more obvious.

\section{Numerical Results}
\label{sec5}

In this section, we discuss the performance of our curvature reconstruction methods on both synthetic and real imaging data by comparing with state-of-the-art methods including the direction methods, i.e., LCT \cite{o2018confocal}, Phasor field \cite{liu2019non} and F-K migration \cite{lindell2019wave}, and iterative methods, i.e., SPIRAL+$\|\cdot\|_1+\mathbb R_+$ (shorted by SPIRAL) \cite{ye2021compressed} and SOCR \cite{liu2021non}. 

\begin{figure*}
	\centering
        \captionsetup[subfloat]{labelsep=none,format=plain,labelformat=empty}
        
    ~~\subfloat[\footnotesize{(a)Ground truth}]{\includegraphics[width=0.9in,height=26mm]{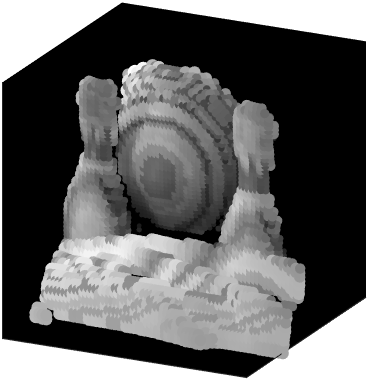}}
    \subfloat[\footnotesize{(b)LCT}]{\includegraphics[width=0.9in,height=26mm]{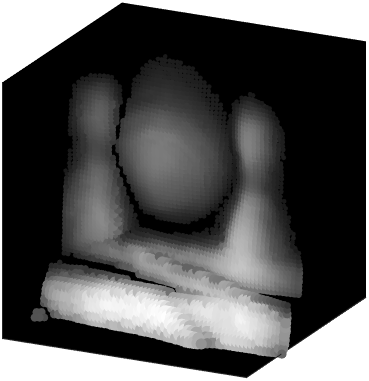}}
    \subfloat[\footnotesize{(c)Phasor field}]{\includegraphics[width=0.9in,height=26mm]{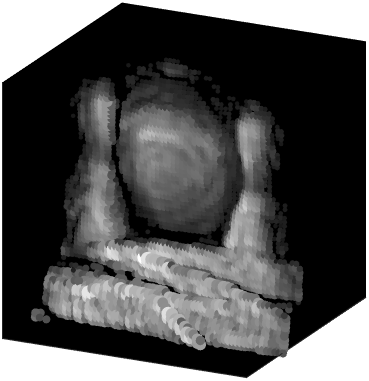}}
	\subfloat[\footnotesize{(d)F-K}]{\includegraphics[width=0.9in,height=26mm]{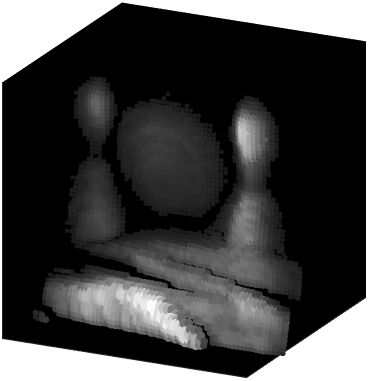}}
     \subfloat[\footnotesize{(e)SOCR}]{\includegraphics[width=0.9in,height=26mm]{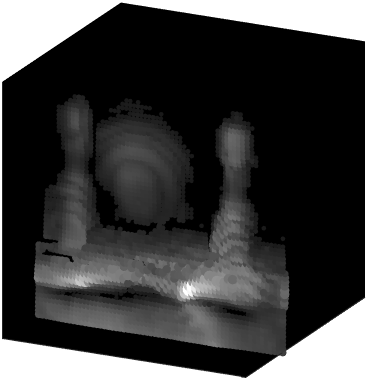}}
       \subfloat[\footnotesize{(f)SPIRAL}]{\includegraphics[width=0.9in,height=26mm]{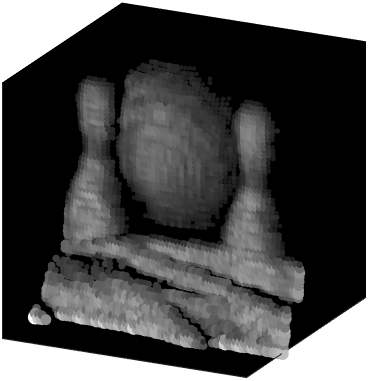}}
    \subfloat[\footnotesize{(g)Algorithm \ref{alg1}}]{\includegraphics[width=0.9in,height=26mm]{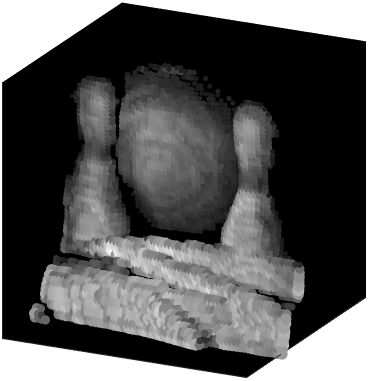}}
    \subfloat[\footnotesize{(h)Algorithm \ref{alg2}}]{\includegraphics[width=0.9in,height=26mm]{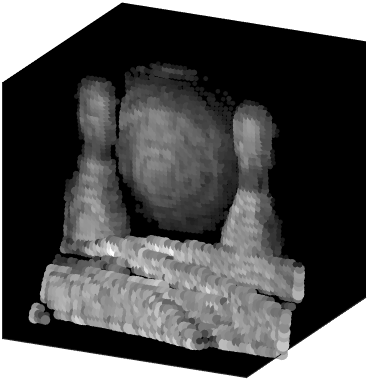}}\\
    \vspace{0.02cm}
    \captionsetup{font = small}
    \caption{The visual comparison of the comparison reconstruction methods under full sampling on Bowling, where the parameters of our methods are set as: $a = 5\times10^{-5}, b = 5\times10^{-5}$, $\mu$ = 0.1 for Algorithm \ref{alg1}; $a_{\tau} = 1\times10^{-4}, b_{\tau} = 3\times10^{-2},a_u = 1\times10^{-4}, b_u = 1\times10^{-4}, \lambda = 400$ for Algorithm \ref{alg2}.}
 \label{fig:bowling full samping}
\end{figure*}

\subsection{Parameter discussing}
For the object-domain Algorithm \ref{alg1}, three parameters need to be adjusted, namely $\mu, a, b$. The penalty parameter $\mu$ controls the convergence of the algorithm. Too small $\mu$ will lead to non-convergence of the algorithm, and too large $\mu $ will reduce the quality of the reconstructed image. The two parameters $a$ and $b$ are the regularization parameters used to control the smoothness of the solution. The larger the values of $a$ and $b$ are, the smoother the surfaces will be. The specific values of the three parameters are provided in each scene. There are total eight parameters $\lambda$, $\mu_1, \mu_2, \mu_3, a_u, b_u, a_{\tau }, b_{\tau}$ in the dual-domain Algorithm \ref{alg2}. Similarly, the penalty parameters $\mu_1, \mu_2, \mu_3$ affect the stability of the algorithm, which are fixed $\mu_1=1,\mu_2=800,\mu_3=2$ in all experiments. Varying these values in a small range will not affect the quality of the reconstructed image. The other five arguments are used to balance the data fidelity and curvature regularity. More specifically, $a_{u}$ and $b_{u}$ control the curvature regularization of the object domain, while $a_{\tau}$ and $b_{\tau}$ control the curvature regularization of the measured signals. Likewise, we provide the specific values in each experiment. In addition, both Algorithm \ref{alg1} and Algorithm \ref{alg2} are terminated by both the number of iterations and the relative error bound. Because Algorithm \ref{alg1} converges faster than Algorithm \ref{alg2}, the number of iterations of Algorithm \ref{alg1} is set to $T_{max}=200$, and the number of iterations of Algorithm \ref{alg2} is set to $T_{max}=300$. To ensure the quality of the reconstructed image, we set the relative error bound $\epsilon$ as $1\times 10^{-6}$.

\subsection{Experiments on synthetic data}
In this subsection, we use two synthetic confocal data and one non-confocal data to verify the reconstruction performance of our curvNLOS methods.

\begin{table}
	\caption{The comparison of the Accuracy, RMSE, PSNR and SSIM among different methods on Bowling with scanning points of $64\times64$, $8\times8$, $6\times6$, $4\times4$.}
	\label{table:multi-samp}
	\centering
		\begin{tabular}{c|c|c|c|c|c}
			\Xhline{1.5pt}
			{Scan points}&{Method}&{Accuracy}&{RMSE}&{PSNR}&{SSIM}\\
			\hline
             \multirow{7}{*}{64$\times$64}
                &LCT&0.8818 & 0.1977 &  13.9723 & 0.2385\\
			\cline{2-6}
			& Phasor field& 0.9333 & 0.1712 &  13.9628 & 0.2344\\
			\cline{2-6}
			&F-K&0.8594 & 0.2055 & 10.9895  & 0.1826\\
                \cline{2-6}
                &SOCR&0.8889 & 0.5770  & 11.1099  & 0.2311\\
                \cline{2-6}
			&SPIRAL& 0.9445& 0.1391 &  13.7114 &0.4346 \\
			\cline{2-6}
			&Algorithm \ref{alg1}&0.9558 &0.1070  & 16.2712  &0.4473  \\
			\cline{2-6}
			&Algorithm \ref{alg2}&0.9585 & 0.0946 &  16.1188 &0.4758  \\
			\hline
			\multirow{3}{*}{8$\times$8}
			&SPIRAL& 0.9177 & 0.1606 &  13.2804 &0.2589 \\
			\cline{2-6}
			&Algorithm \ref{alg1}& 0.9236 & 0.1517 & 14.7608  & 0.3098 \\
			\cline{2-6}
			&Algorithm \ref{alg2}& 0.9431 & 0.1461 & 15.3378  & 0.3156 \\
			\hline
            \multirow{3}{*}{6$\times$6}
			&SPIRAL& 0.9067 &0.1652  & 13.1770  &0.2119 \\
			\cline{2-6}
			&Algorithm \ref{alg1} & 0.9226 & 0.1513 & 14.6333 & 0.2742  \\
			\cline{2-6}
			&Algorithm \ref{alg2}& 0.9358 & 0.1488 & 14.7758  & 0.2924\\
			\cline{2-6}
			\hline
			\multirow{3}{*}{4$\times$4}
			&SPIRAL& 0.8621 & 0.2087 & 12.4577 & 0.1664 \\
			\cline{2-6}
			&Algorithm \ref{alg1}& 0.8782 & 0.1982 &  14.5064 & 0.2220 \\
			\cline{2-6}
			&Algorithm \ref{alg2}& 0.8977 & 0.1948 &  14.5900 &0.2304 \\
			\cline{2-6}
			\Xhline{1.5pt}
		\end{tabular}
\end{table}

\begin{figure}
	\centering
    \captionsetup[subfloat]{labelsep=none,format=plain,labelformat=empty}
	\rotatebox{90}{\scriptsize{\qquad\quad\textbf{8$\times$8}}}
    \subfloat{\includegraphics[width=0.67in,height=20mm]{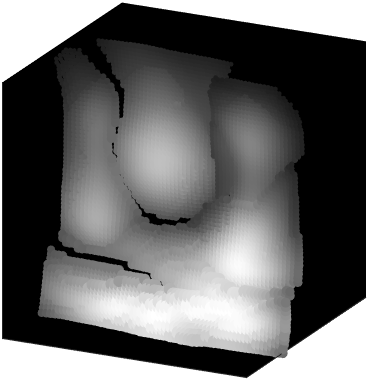}}
    \subfloat{\includegraphics[width=0.67in,height=20mm]{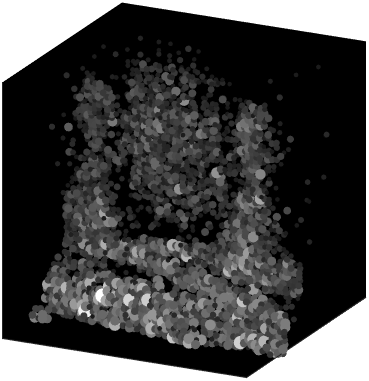}}
    \subfloat{\includegraphics[width=0.67in,height=20mm]{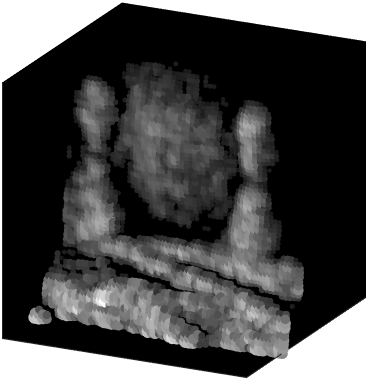}}
	\subfloat{\includegraphics[width=0.67in,height=20mm]{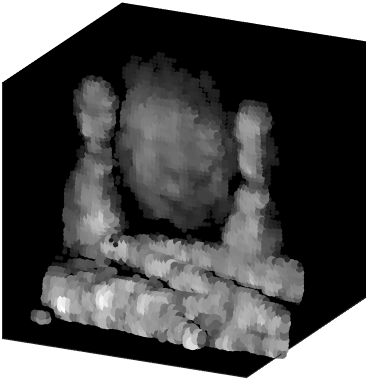}}
	\subfloat{\includegraphics[width=0.67in,height=20mm]{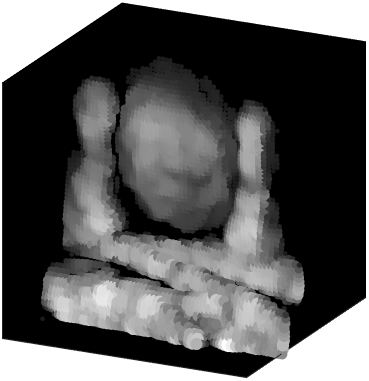}}\\
    \vspace{0.02cm}
	\rotatebox{90}{\scriptsize{\qquad\quad\textbf{6$\times$6}}}
    \subfloat{\includegraphics[width=0.67in,height=20mm]{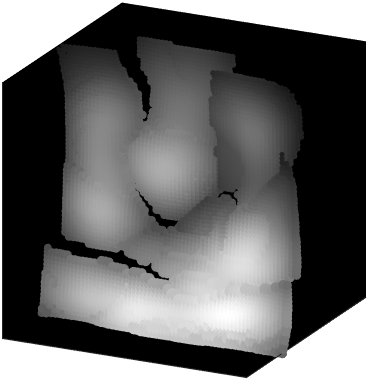}}
    \subfloat{\includegraphics[width=0.67in,height=20mm]{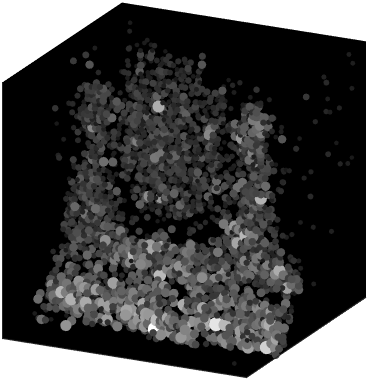}}
    \subfloat{\includegraphics[width=0.67in,height=20mm]{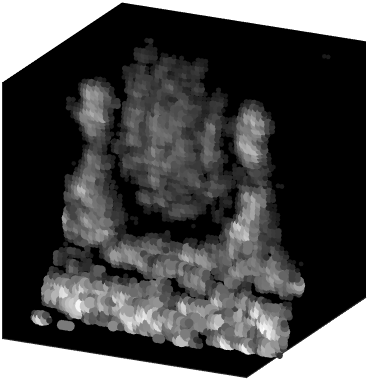}}
	\subfloat{\includegraphics[width=0.67in,height=20mm]{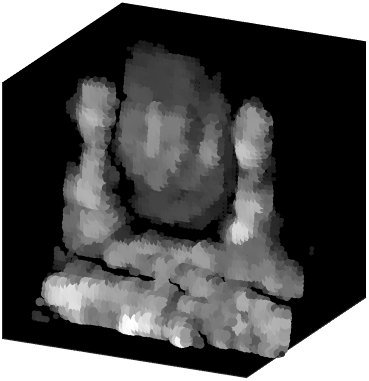}}
	\subfloat{\includegraphics[width=0.67in,height=20mm]{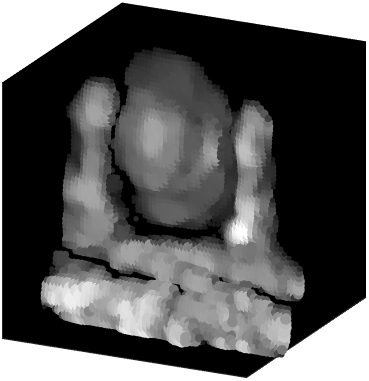}}\\
        \vspace{0.02cm}
	\setcounter{subfigure}{0}
	\rotatebox{90}{\scriptsize{\qquad\quad\textbf{4$\times$4}}}
    \subfloat[(a)]{\includegraphics[width=0.67in,height=20mm]{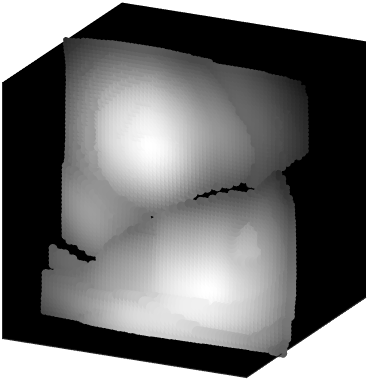}}
	\subfloat[(b)]{\includegraphics[width=0.67in,height=20mm]{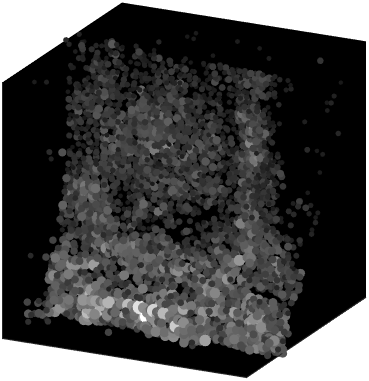}}
	\subfloat[(c)]{\includegraphics[width=0.67in,height=20mm]{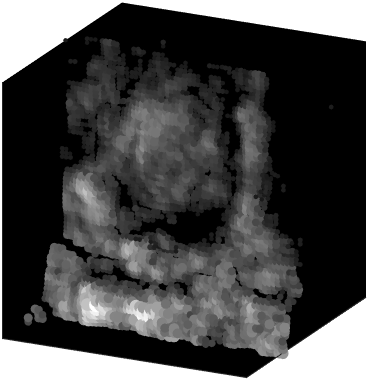}}
	\subfloat[(d)]{\includegraphics[width=0.67in,height=20mm]{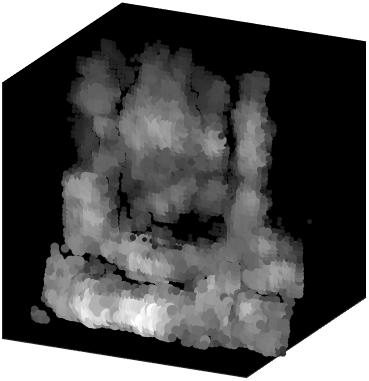}}
	\subfloat[(e)]{\includegraphics[width=0.67in,height=20mm]{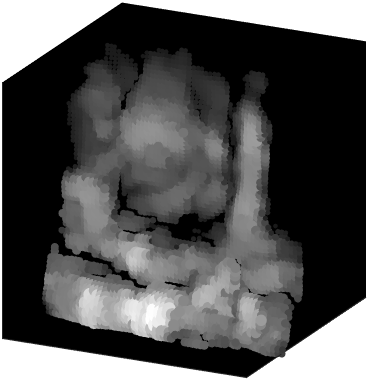}}
    \captionsetup{font = small}
    \caption{Reconstruction results using different numbers of scanning points from up to down the scanning points are of $8\times 8$, $6\times 6$, and $4\times4$, respectively, where (a) LCT, (b) SPIRAL w/o smoothing, (c) SPIRAL with smoothing, (d) Algorithm 1 and (e) Algorithm 2. The parameters of our methods are set as: $a = 9\times10^{-5}, b = 2\times10^{-5}$ ($8\times8$ ), $a = 1\times10^{-6}, b = 4\times10^{-5}$ ($6\times6$ ), $a = 1\times10^{-5}, b = 1\times10^{-5}$ ($4\times4$) and $\mu$ = 0.1 for Algorithm \ref{alg1}; $\lambda = 235, a_{\tau} = 4\times10^{-4}, b_{\tau} = 3\times10^{-2},a_u = 2\times10^{-4}, b_u = 4.5\times10^{-4}$ ($8\times8$), $\lambda = 270, a_{\tau} = 2\times10^{-4}, b_{\tau} = 4\times10^{-2},a_u = 8\times10^{-4}, b_u = 1\times10^{-4}$ ($6\times6$), $\lambda = 300,a_{\tau} = 1\times10^{-4}, b_{\tau} = 1.2\times10^{-2},a_u = 6\times10^{-4}, b_u = 3\times10^{-4}$($4\times4$) for Algorithm \ref{alg2}.}
	\label{fig:multi-samp visual comparison bowling}
\end{figure}

\subsubsection{Results on Bowling w.r.t. different scanning points}
The bowling scene was generated in \cite{ye2021compressed}, where 64 $\times$64 scanning points are used to cover a 1m$\times$1m square on the wall with a temporal resolution of 256 with bins of width $32$ ps.   Firstly, we compare our reconstruction results with other methods using the full sampling data; see Fig. \ref{fig:bowling full samping} for a visual comparison. 
As can be observed in Fig. \ref{fig:bowling full samping}, different methods can produce meaningful reconstruction results on the full sampling data. Our reconstructions achieve the best visual quality, where the reconstructed scenes are quite close to the ground truth with fine structures and details. Table \ref{table:multi-samp} records the evaluation indicators including Accuracy, RMSE, PSNR, and SSIM for all comparison methods, where our curvNLOS gives the best accuracy. 
Although the object-domain reconstruction (Algorithm \ref{alg1}) and dual-domain reconstruction (Algorithm \ref{alg2}) provide similar visual results, the quantitative indexes indicate Algorithm \ref{alg2} provides the best quality.

Secondly, we compare the performance for under-sampled sparse reconstruction problems; see Fig. \ref{fig:multi-samp visual comparison bowling}. For under-sampled scanning, the measurement data for all comparison methods except for SPIRAL was obtained by linear interpolation filling. 
The first column of Fig. \ref{fig:multi-samp visual comparison bowling} displays the reconstruction results of LCT using $8 \times 8$, $6 \times 6$, and $4 \times 4$ scanning points. It is difficult to identify the meaningful scene information from the reconstructed images. The same is true for the Phasor field and F-K. The second column and the third column of Fig. \ref{fig:multi-samp visual comparison bowling} are the reconstruction results of SPIRAL without smoothing and after smoothing. As can be seen, smoothing plays a very important role for SPIRAL, while our approaches do not involve a smoothing post-processing step. The last two columns are the reconstruction obtained by Algorithm \ref{alg1} and Algorithm \ref{alg2}, respectively. We can observe that the image contrast is significantly improved, and the structural information is much clearer than both LCT and SPIRAL, especially for $4\times4$ scanning points. The Algorithm \ref{alg2} tends to produce reconstruction results with better smoothness due to the introduction of the curvature regularization for measured signals. When the scanning points become more and more sparse, the advantages of the Algorithm \ref{alg2} become more and more obvious. Furthermore, Table \ref{table:multi-samp} exhibits the metrics estimated by SPIRAL with smoothing and our methods, which further convinced the advantages of our methods over SPIRAL. 

\begin{figure}[t]
	\centering
	\begin{minipage}[t]{0.3\linewidth}
		\centering
             \subfloat[\footnotesize SPIRAL]{\includegraphics[width=1.12in,height=28mm]{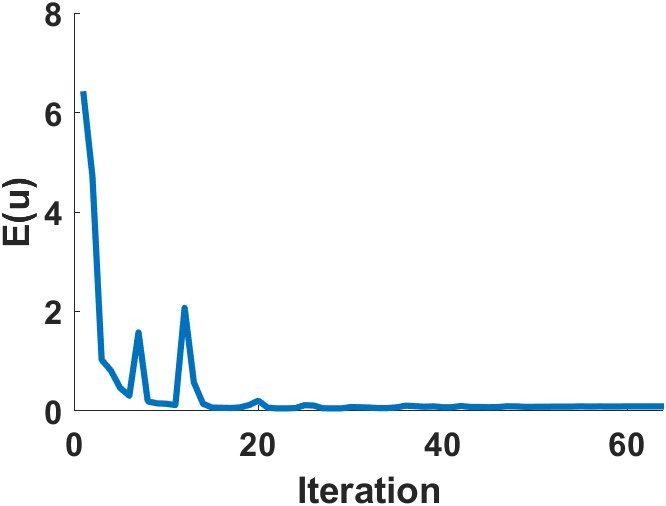}}\hspace{1ex}\\
		\vspace{0.2cm}
	\end{minipage}%
	~~
	\begin{minipage}[t]{0.3\linewidth}
		\centering
            \subfloat[Algorithm 1]{\includegraphics[width=1.12in,height=28mm]{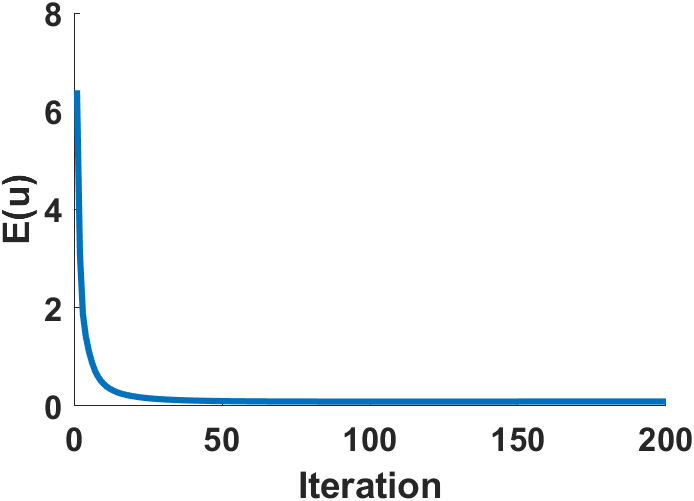}}\hspace{0.5ex}\\
            \vspace{0.02cm}
	\end{minipage}%
	~~	
	\begin{minipage}[t]{0.28\linewidth}
		\centering
	      \subfloat[Algorithm 2] {\includegraphics[width=1.12in,height=28mm]{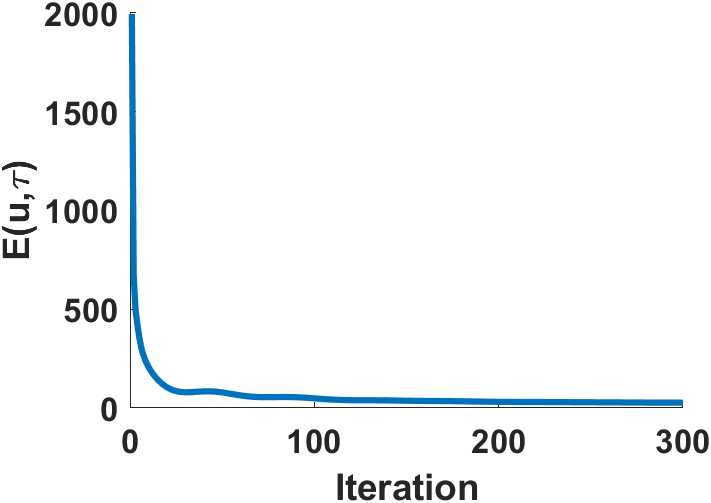}}\hspace{0.5ex}\\
            \end{minipage}
	\centering
	\caption{The comparison of energy decays between SPIRAL in \cite{ye2021compressed} and our Algorithm \ref{alg1}, Algorithm \ref{alg2}.}
	\label{curve}
\end{figure}

\subsubsection{The effect of curvature function $\phi$}
We assess the impact of the curvature function on the quality of reconstruction performance. We compared the reconstruction results with different curvature functions on bowling data of $6\times 6$ scanning points. As shown in Table \ref{table:curvature}, Algorithms 1 and Algorithm 2 achieve the best reconstruction accuracy when using TSC as the curvature function. Therefore, we use TSC as the curvature function in both the object-domain model \eqref{image_model} and the dual-domain model \eqref{dual_model} for the rest of the numerical experiments.

\begin{table}[!htbp]
	\caption{The comparison of different functionals of curvature in terms of Accuracy, RMSE, PSNR, and SSIM on $6\times6$ bowling data.}
	\label{table:curvature}
	\centering
		\begin{tabular}{c|c|c|c|c|c}
			\Xhline{1.5pt}
{Method}&$\phi$&{Accuracy}&{RMSE}&{PSNR}&{SSIM}\\
			\hline
\multirow{3}{*}{Algorithm 1} 
			&TRV& 0.9177 & 0.1583  & 14.5077  &\bf{0.2953} \\
			\cline{2-6}
			&TAC& \bf{0.9243} & \bf{0.1500} &  13.4762 &0.2721 \\
			\cline{2-6}
			&TSC & 0.9226 &0.1513  & \bf{14.6333}  &0.2742 \\
			\cline{1-6}
			\multirow{3}{*}{Algorithm 2} 
			&TRV& 0.9290 & 0.1548 &  \bf{14.8599} &\bf{0.3094} \\
			\cline{2-6}
			&TAC& 0.9290 & 0.1528 &  14.4663 &0.2875 \\
			\cline{2-6}
			&TSC& \bf{0.9358} & \bf{0.1488} &  14.7758 &0.2924 \\
			\Xhline{1.5pt}
		\end{tabular}
\end{table}

\subsubsection{Numerical convergence}
In what follows, we examine the energy decay of the SPIRAL, our Algorithm 1, and Algorithm 2. As shown in Fig.\ref{curve}, the SPIRAL converges much more quickly than our Algorithm 1 and Algorithm 2. However, its calculation requires inner iteration, resulting in the fluctuation of the numerical energies in the early stage. On the other hand, our algorithms demonstrate stable energy convergence and satisfactory numerical performance through the iteration process.
 \begin{figure}[h]
	\centering
	\begin{minipage}[t]{0.3\linewidth}
		\centering
   \subfloat{\includegraphics[width=1in,height=28mm]{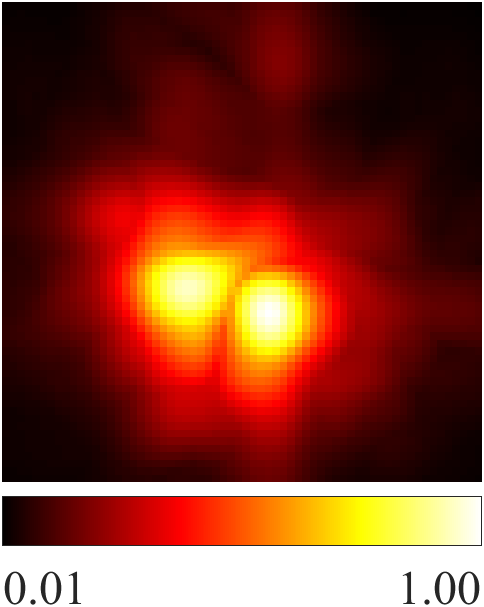}}\\
\subfloat{\includegraphics[width=1in,height=28mm]{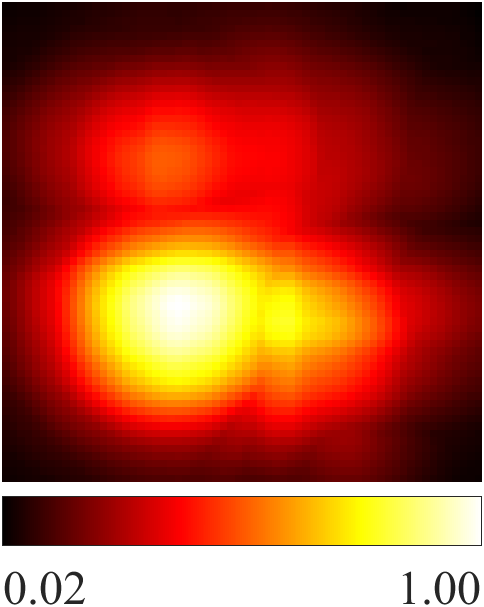}}\\
             {\small{(a)LCT}}
	\end{minipage}%
	\begin{minipage}[t]{0.3\linewidth}
		\centering
            \subfloat{\includegraphics[width=1in,height=28mm]{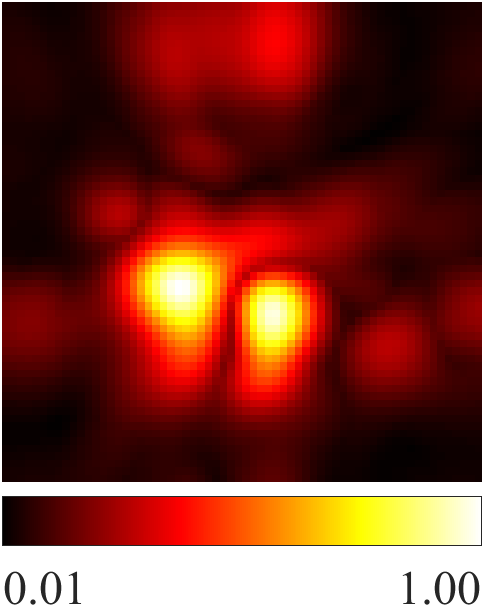}}\\
            \subfloat{\includegraphics[width=1in,height=28mm]{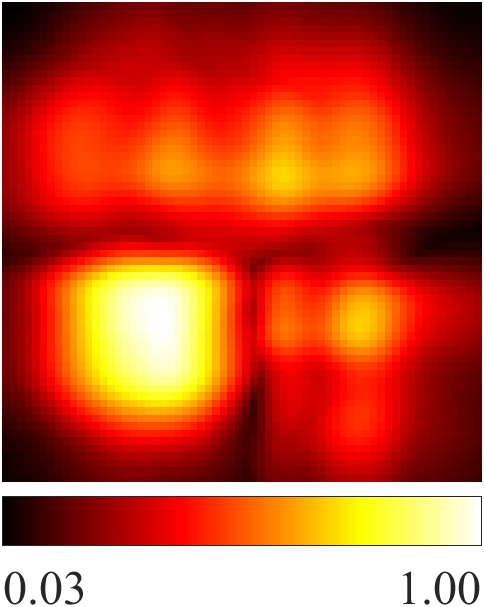}}\\
            {\small{(b)Phasor field}}
	\end{minipage}%
	\begin{minipage}[t]{0.3\linewidth}
		\centering
            \subfloat{\includegraphics[width=1in,height=28mm]{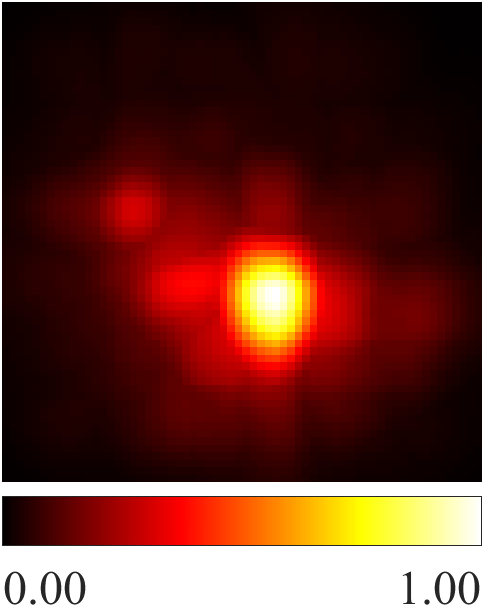}}\\
            \subfloat{\includegraphics[width=1in,height=28mm]{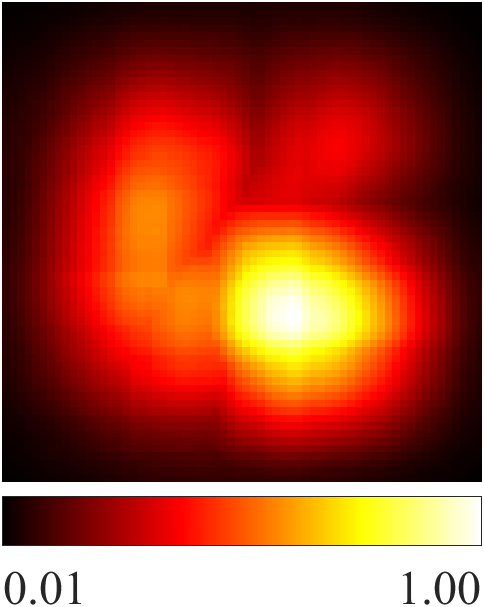}}\\
            {\small{(c)F-K}}
	\end{minipage}
	\centering
	\caption{Comparison for the reconstruction results of under-sampled Stanford bunny, where scanning points are of resolution $8\times8$ and $4\times4$, respectively.}
        \label{fig:bunny undersampleddirect}
\end{figure}

\begin{figure}
	\centering
    \captionsetup[subfloat]{labelsep=none,format=plain,labelformat=empty}
	\rotatebox{90}{\scriptsize{\qquad\qquad\quad\textbf{64$\times$64}}}
    \subfloat{\includegraphics[width=0.8in,height=28mm]{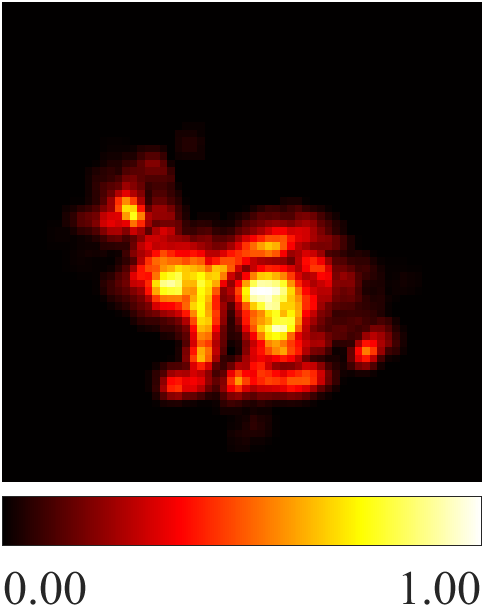}}
    \subfloat{\includegraphics[width=0.8in,height=28mm]{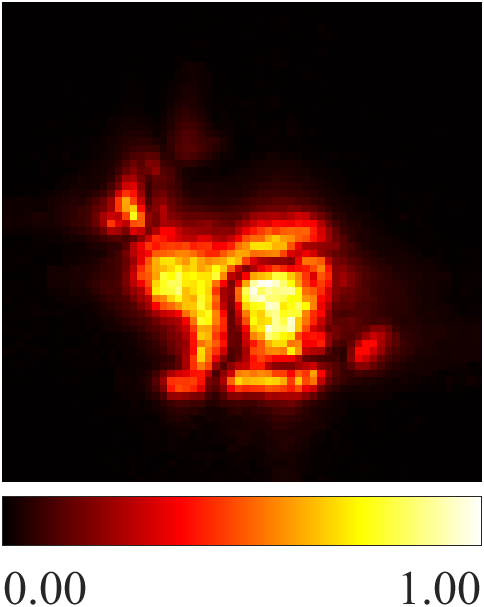}}
	\subfloat{\includegraphics[width=0.8in,height=28mm]{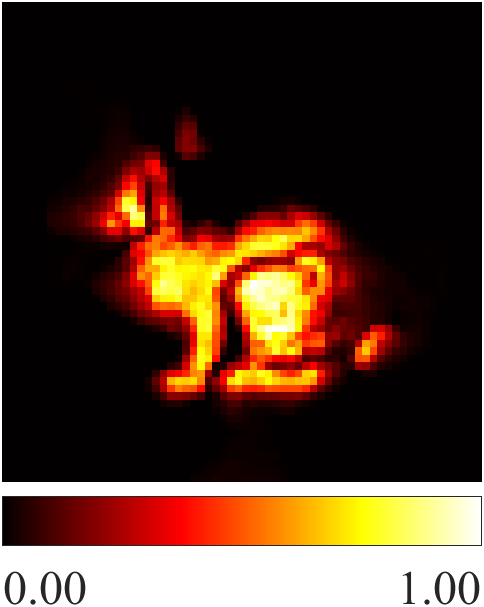}}
	\subfloat{\includegraphics[width=0.8in,height=28mm]{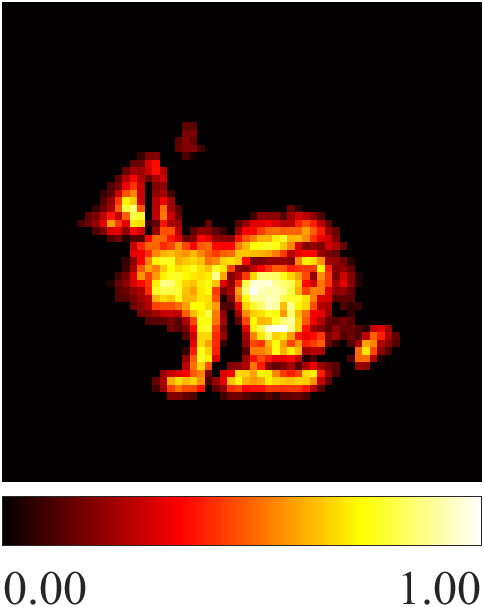}}\\
    \vspace{-0.1in}
    \rotatebox{90}{\scriptsize{\qquad\qquad\quad\textbf{8$\times$8}}}
    \subfloat{\includegraphics[width=0.8in,height=28mm]{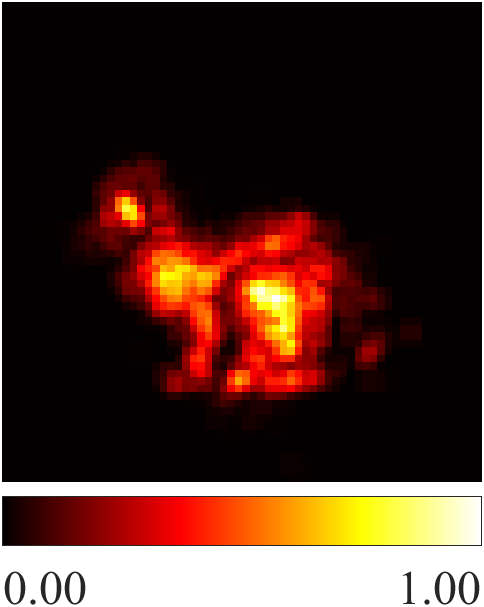}}
    \subfloat{\includegraphics[width=0.8in,height=28mm]{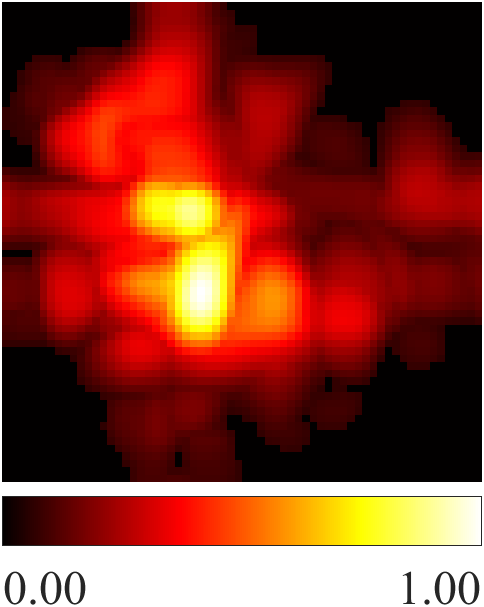}}
	\subfloat{\includegraphics[width=0.8in,height=28mm]{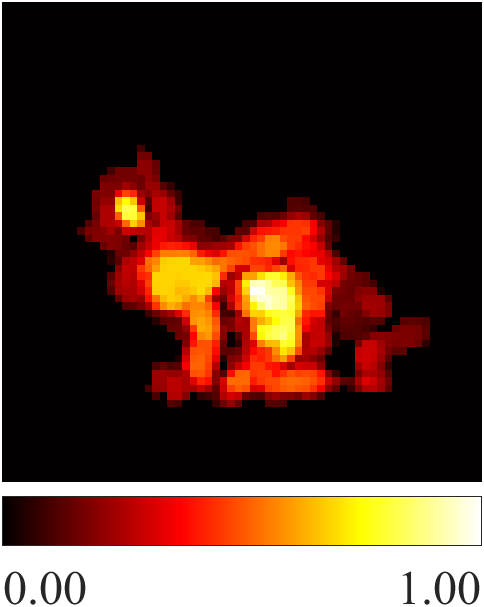}}
	\subfloat{\includegraphics[width=0.8in,height=28mm]{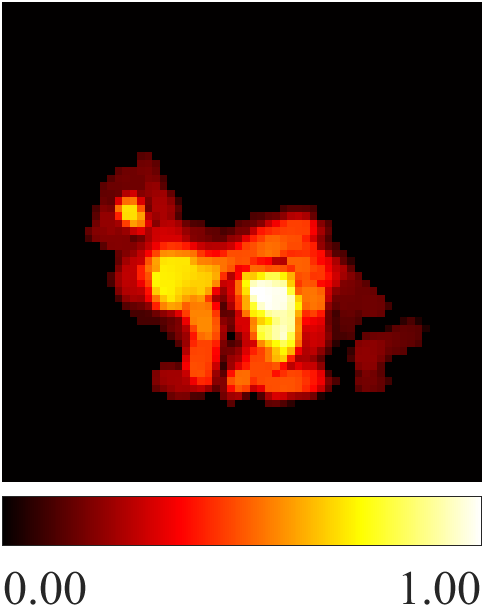}}\\
    \vspace{-0.1in}
	\setcounter{subfigure}{0}
	\rotatebox{90}{\scriptsize{\qquad\qquad\qquad\textbf{4$\times$4}}}
    \subfloat[\footnotesize {(a)SPIRAL} ]{\includegraphics[width=0.8in,height=28mm]{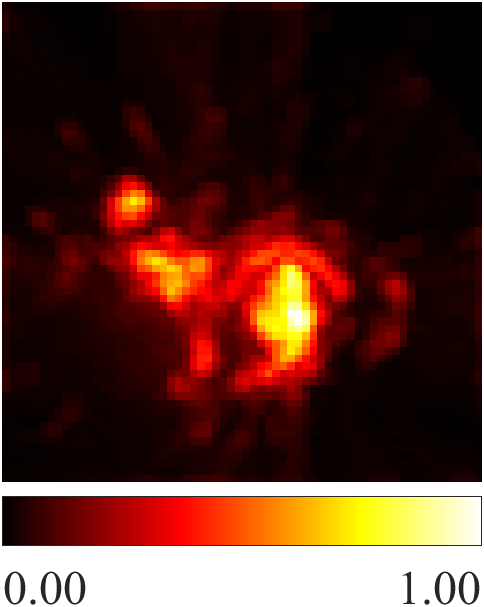}}
	\subfloat[\footnotesize{(b)SOCR}]{\includegraphics[width=0.8in,height=28mm]{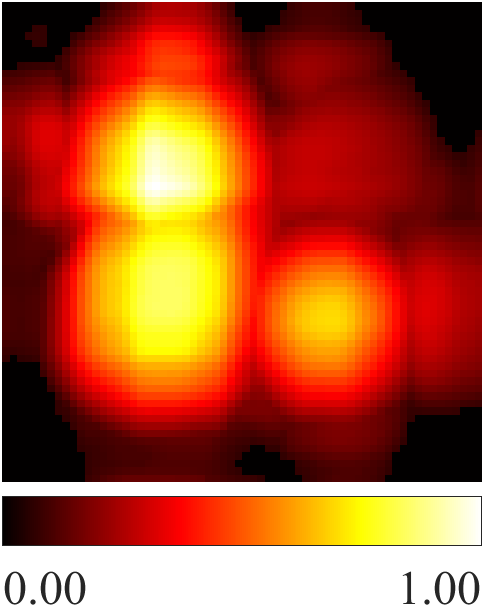}}
	\subfloat[\footnotesize{(c)Algorithm \ref{alg1} }]{\includegraphics[width=0.8in,height=28mm]{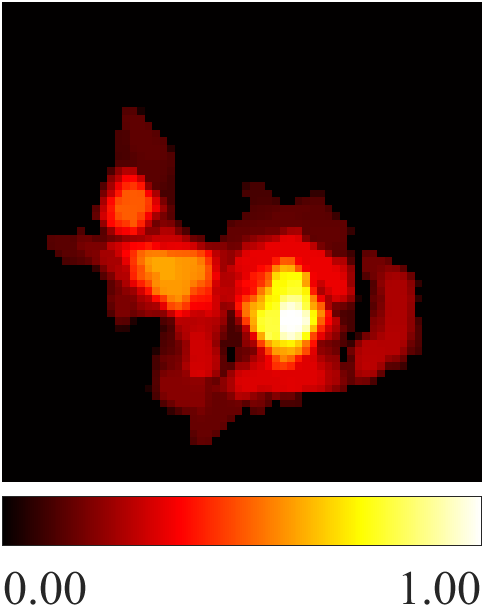}}
	\subfloat[\footnotesize{(d)Algorithm \ref{alg2} }]{\includegraphics[width=0.8in,height=28mm]{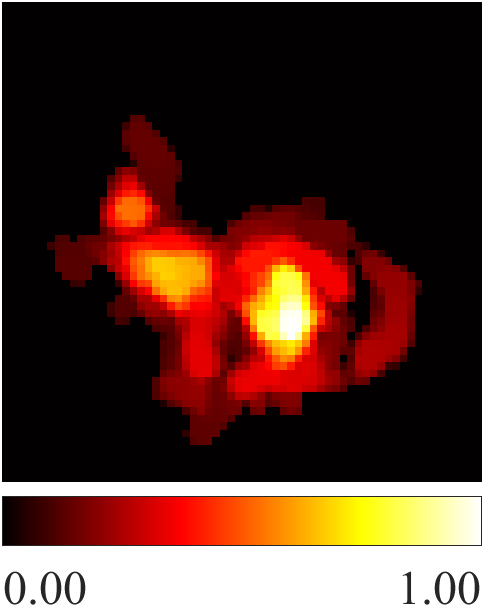}}
    \captionsetup{font = small}
    \caption{Comparison for the reconstruction results of Stanford bunny, where scanning points are of resolution $64\times64$, $8\times8$, and $4\times4$ respectively. The parameters are set as: $a = 1\times10^{-4}, b = 1\times10^{-4}$ ($64\times64$), $a = 1\times10^{-4}, b = 4\times10^{-5}$ ($8\times8$), $a = 1\times10^{-4}, b = 1.5\times10^{-5}$ ($4\times4$) for Algorithm \ref{alg1}; $\lambda = 300,a_{\tau} = 1\times10^{-3}, b_{\tau} = 1\times10^{-3}, a_u = 1\times10^{-4}, b_u = 1\times10^{-5}$ ($64\times64$), $\lambda = 150,a_{\tau} = 5\times10^{-3}, b_{\tau} = 1\times10^{-3},a_u = 5\times10^{-4}, b_u = 3\times10^{-4}$ ($8\times8$), $\lambda = 250,a_{\tau} = 1\times10^{-3}, b_{\tau} = 1\times10^{-3},a_u = 5\times10^{-5}, b_u = 2\times10^{-4}$ ($4\times4$) for Algorithm \ref{alg2}.}
        \label{fig:bunny undersampled}
\end{figure}

\subsubsection{Results on Stanford bunny w.r.t. different scanning points}
Another synthetic data is Stanford bunny from the Zaragoza NLOS synthetic dataset. The total $64\times64$ scanning points occupy an area of $0.6\times0.6$ m$^2$ on the wall. The data has 512 time bins and the photon travels 0.0025 m in each bin. 
As shown in Fig. \ref{fig:bunny undersampleddirect}, when the number of scanning points becomes less than $8\times8$, the quality of images reconstructed by direct methods tends to degrade. 
Fig. \ref{fig:bunny undersampled} shows the comparison results of the iterative methods with different sparse scanning points, where both SPIRAL and our methods can estimate the shape of the bunny for only $4\times 4$ scanning points. 
Our curvNLOS gives better reconstruction quality with much smoother surfaces and fewer artifacts. Although we use the interpolated measurement data for SOCR, it still fails to obtain meaningful reconstruction results, which reveals its limitation in dealing with compressed sensing reconstruction scenarios. 

\subsection{Experiments on measured data}
\begin{figure}
	\centering
        \captionsetup[subfloat]{labelsep=none,format=plain,labelformat=empty}
        \rotatebox{90}{\normalsize{\qquad~ \textbf{$64\times64$}}}
    \subfloat{\includegraphics[width=0.823in,height=25mm]{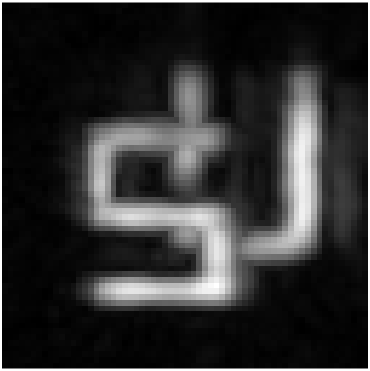}}
       \subfloat{\includegraphics[width=0.823in,height=25mm]{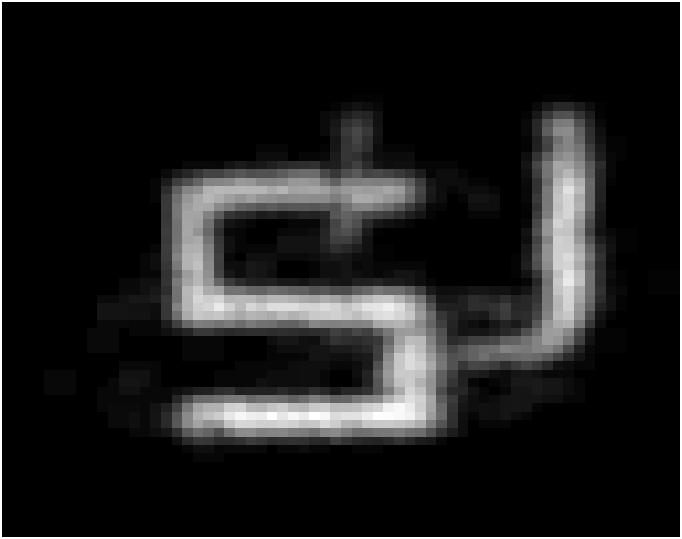}}
    \subfloat{\includegraphics[width=0.823in,height=25mm]{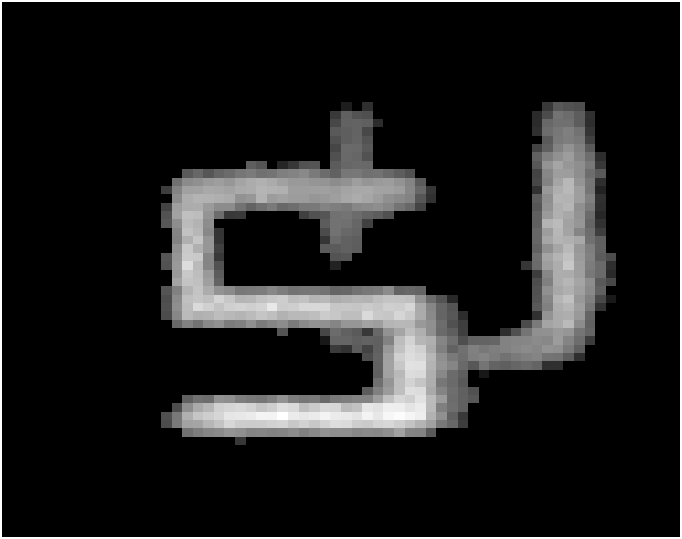}}
    \subfloat{\includegraphics[width=0.823in,height=25mm]{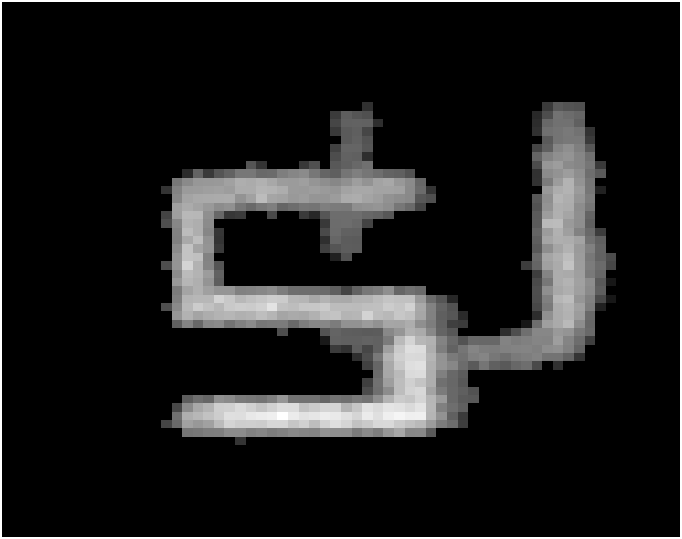}}\\
    \vspace{-0.1in}

    \rotatebox{90}{\normalsize{\qquad\quad \textbf{$8\times8$}}}
    \subfloat{\includegraphics[width=0.823in,height=25mm]{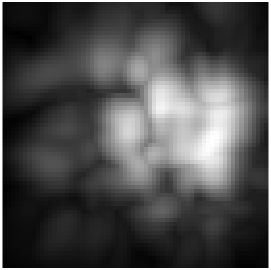}}
    \subfloat{\includegraphics[width=0.823in,height=25mm]{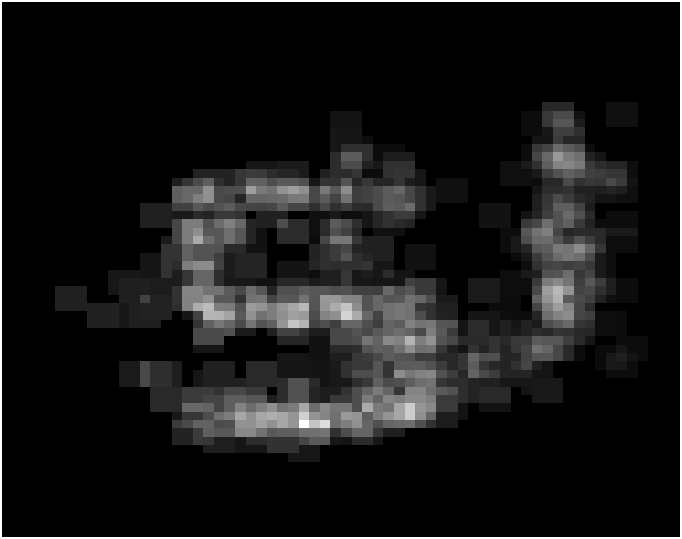}}
    \subfloat{\includegraphics[width=0.823in,height=25mm]{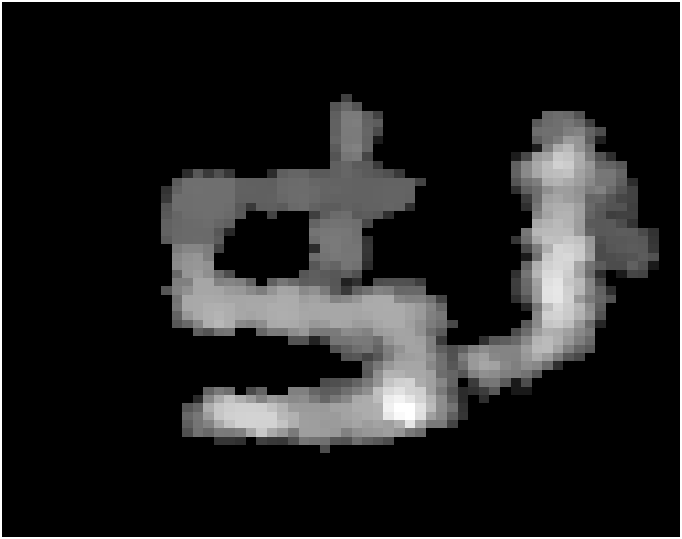}}
    \subfloat{\includegraphics[width=0.823in,height=25mm]{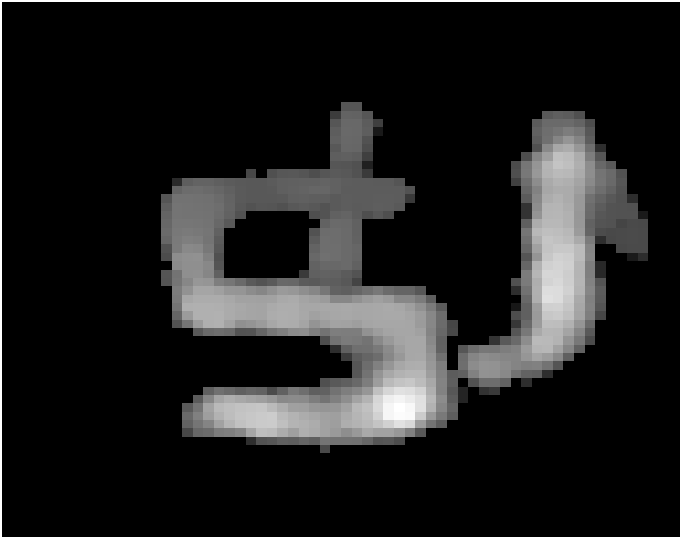}}\\
     \vspace{-0.1in}
    \rotatebox{90}{\normalsize{\qquad\quad~\textbf{$4\times4$}}}
    \subfloat[\footnotesize{(a) Phasor field}]{\includegraphics[width=0.823in,height=28mm]{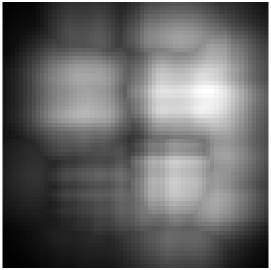}}
    \subfloat[\footnotesize{(b)SPIRAL}]{\includegraphics[width=0.823in,height=28mm]{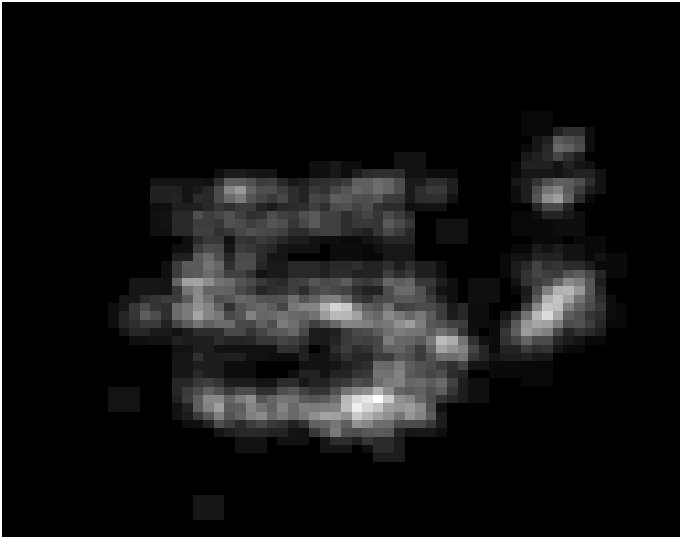}}
    \subfloat[\footnotesize{(c) Algorithm \ref{alg1}}]{\includegraphics[width=0.823in,height=28mm]{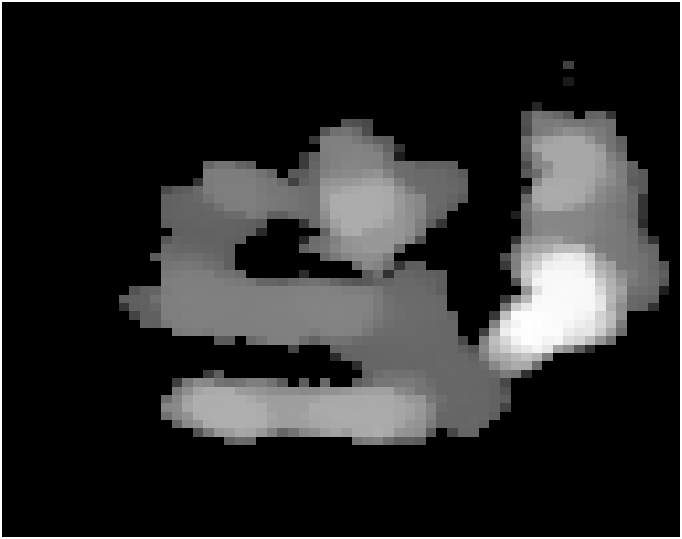}}
    \subfloat[\footnotesize{(d) Algorithm \ref{alg2}}]{\includegraphics[width=0.823in,height=28mm]{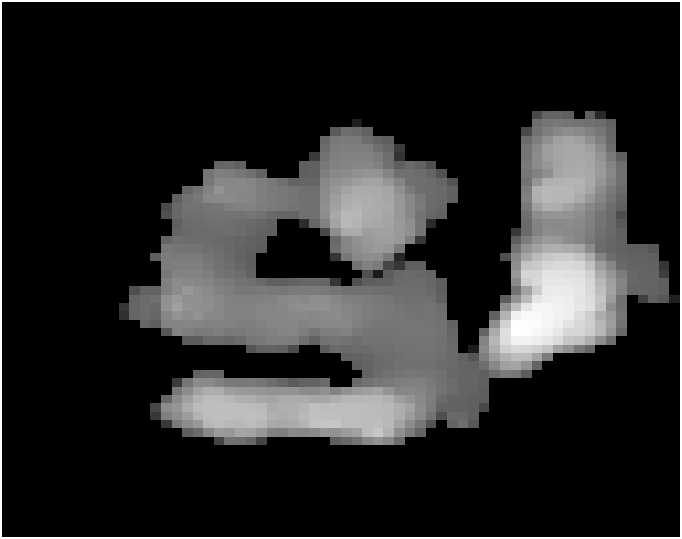}}\\
    \vspace{0.02cm}
    \captionsetup{font = small}
    \caption{ Comparison for reconstruction results of the SU scene, where the parameters are set as: $a = 1.5\times10^{-4}, b = 4\times10^{-5}$ ($64\times64$), $a = 6\times10^{-5}, b = 3\times10^{-5}$ ($8\times8$), $a = 7\times10^{-5}, b = 1.5\times10^{-5}$ ($4\times4$) for Algorithm \ref{alg1}; $\lambda = 200,~a_{\tau} = 3\times10^{-3}, ~b_{\tau} = 5\times10^{-4},~a_u = 5\times10^{-4},~b_u = 5\times10^{-5}$ ($64\times64$), $\lambda = 100,a_{\tau} = 3\times10^{-3}, b_{\tau} = 3\times10^{-3},a_u = 8\times10^{-4}, b_u = 3\times10^{-5}$ ($8\times8$), $\lambda = 150,a_{\tau} = 1\times10^{-3}, b_{\tau} = 1\times10^{-3},a_u = 3\times10^{-4}, b_u = 1\times10^{-5}$ ($4\times4$) for Algorithm \ref{alg2}.}
	\label{fig:su}
\end{figure}

To further prove the effectiveness of our curvNLOS methods, we evaluate them on measured data of the real scenes in \cite{o2018confocal} and \cite{lindell2019wave}. The scene "SU" consists of two letter planes, with the front 'S' obscuring the back 'U', which was sampled at $64\times64$ locations on the wall of size $0.7 \times 0.7$ m$^2$. The time resolution is 512 and each time bin spans $16$ ps. We verify our methods using different sampling rates in the "SU" scene, the results of which are shown in Fig. \ref{fig:su}. 
Due to the severe degradation of the direct methods, we use the Phasor field as a representative of the under-sampled data. As can be observed, our curvNLOS methods can produce satisfactory reconstruction results even when the number of scanning points is reduced to $4\times4$. Compared to SPIRAL, our results preserve the structure of two letters, making them visually clearer and more continuous.

\begin{figure*}[t]
	\centering
        \captionsetup[subfloat]{labelsep=none,format=plain,labelformat=empty}
    \subfloat[\footnotesize{(a) Ground truth}]{\includegraphics[width=0.855in,height=28mm]{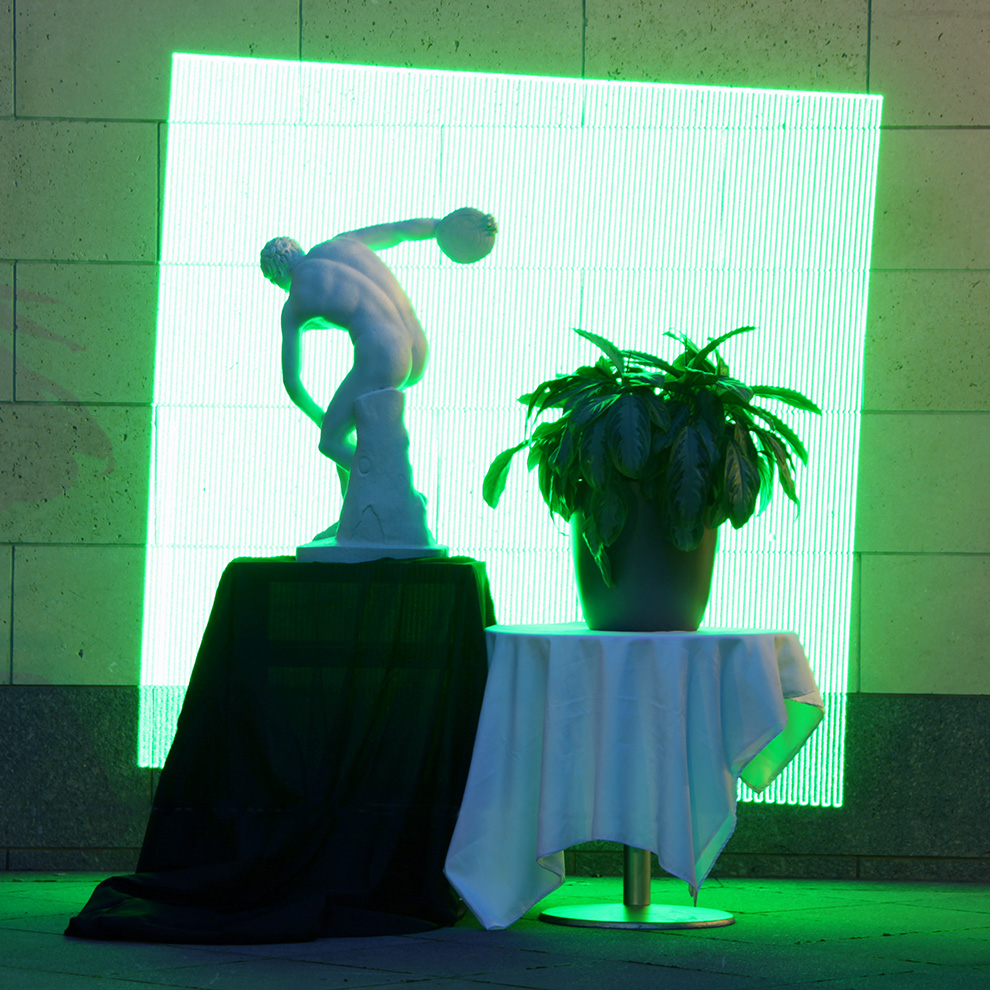}}\hspace{-0.3ex}
    \subfloat[\footnotesize{ (b) LCT}] {\includegraphics[width=0.855in,height=28mm]{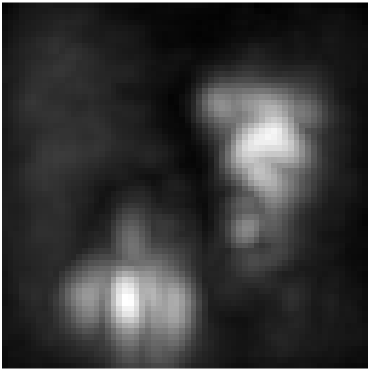}}\hspace{-0.3ex}
    \subfloat[\footnotesize{(c)Phasor field}] {\includegraphics[width=0.855in,height=28mm]{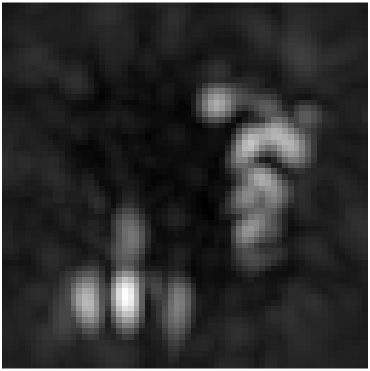}}\hspace{-0.3ex}
    \subfloat[\footnotesize{(d) F-K}]{\includegraphics[width=0.855in,height=28mm]{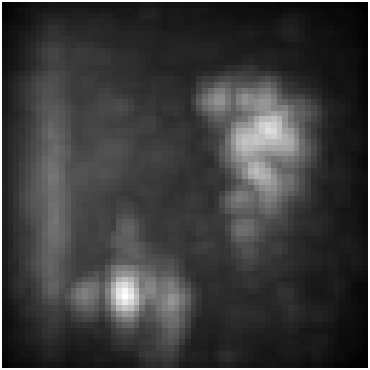}}\hspace{-0.3ex}
   \subfloat[\footnotesize{(e) SPIRAL}]{\includegraphics[width=0.855in,height=28mm]{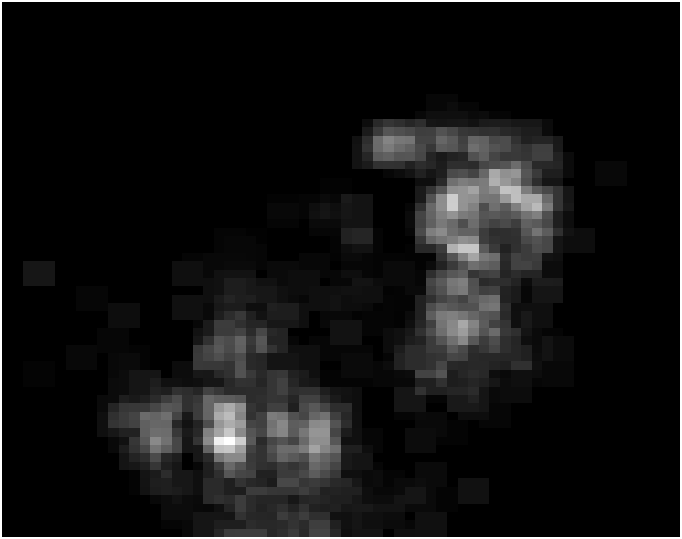}}\hspace{-0.3ex}
   \subfloat[\footnotesize{(f) SOCR}]{\includegraphics[width=0.855in,height=28mm]{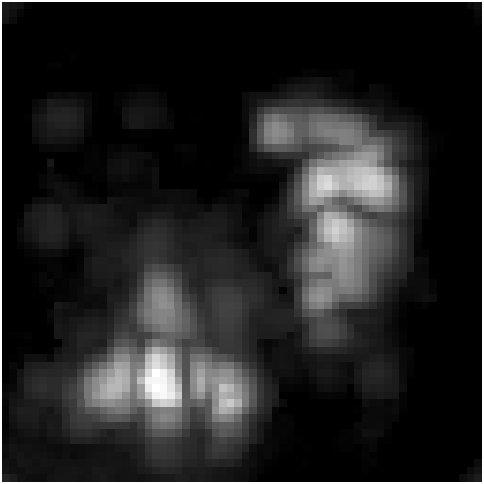}}\hspace{-0.3ex}
   \subfloat[\footnotesize{(g) Algorithm \ref{alg1}}]{\includegraphics[width=0.855in,height=28mm]{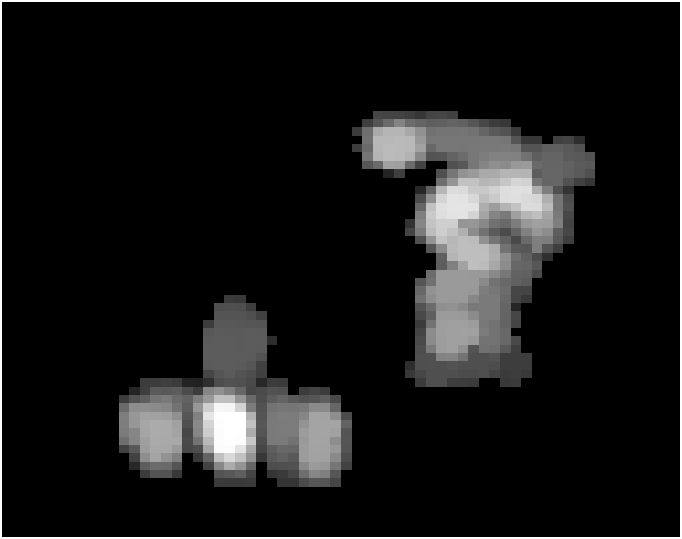}}\hspace{-0.3ex}
   \subfloat[\footnotesize{ (h) Algorithm \ref{alg2}}]{\includegraphics[width=0.855in,height=28mm]{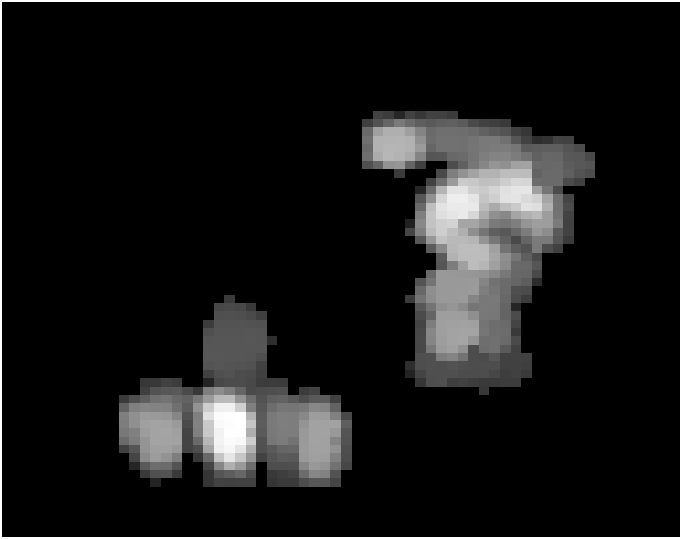}}\hspace{-0.3ex}\\
   \captionsetup{font = small}
	\caption{Comparison for reconstruction results of the outdoor scene (10 min), where the scanning points are of resolution $16\times 16$. The parameters are set as: $\mu$ = 1, $a = 5\times10^{-5}, b = 2\times10^{-4}$ for Algorithm \ref{alg1}; $\lambda = 50,a_{\tau} = 1\times10^{-4}, b_{\tau} = 1\times10^{-4},a_u = 5\times10^{-4}, b_u = 1\times10^{-4}$ for Algorithm \ref{alg2}.}
	\label{fig:outdoor}
\end{figure*}

\begin{figure*}[t]
	\centering
        \captionsetup[subfloat]{labelsep=none,format=plain,labelformat=empty}
    \subfloat[\footnotesize{(a) Ground truth}]{\includegraphics[width=0.98in,height=30mm]{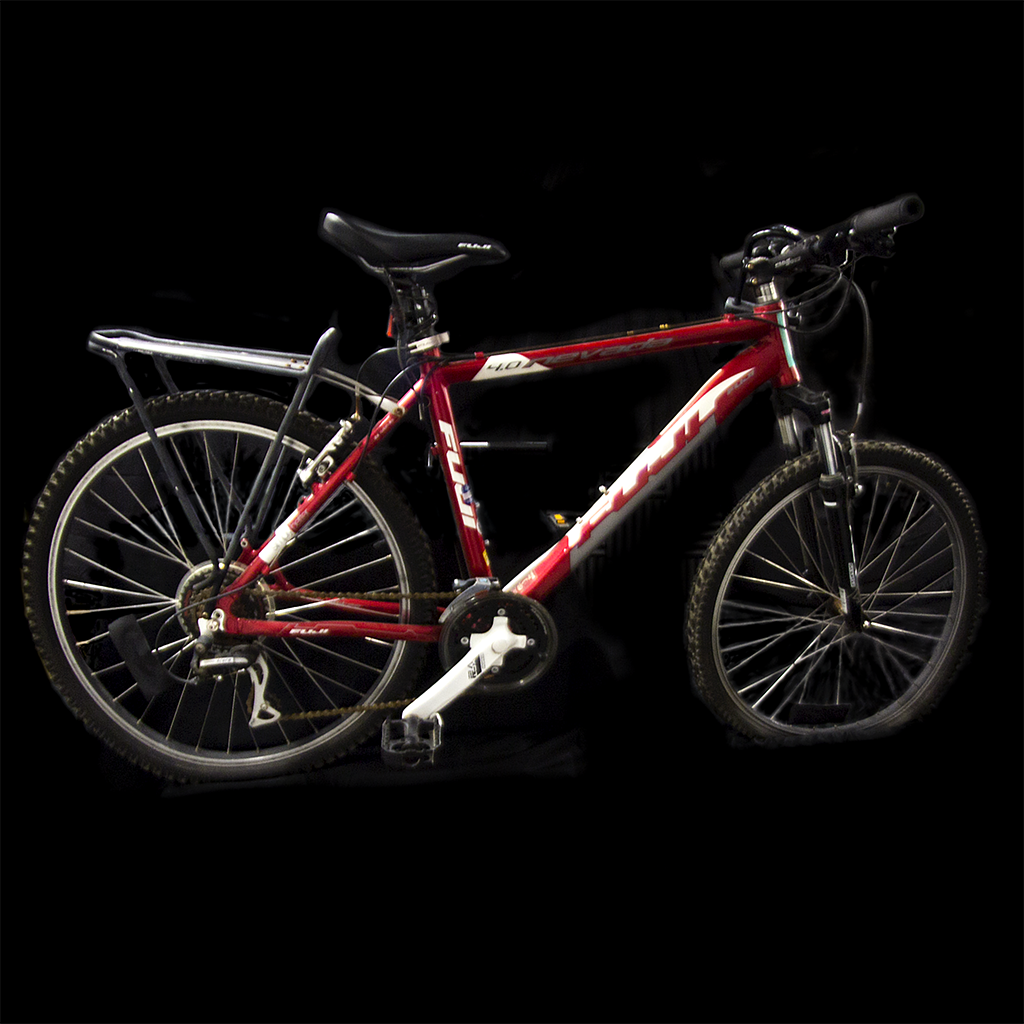}}\hspace{-0.3ex}
    \subfloat[\footnotesize{ (b) LCT}] {\includegraphics[width=0.98in,height=30mm]{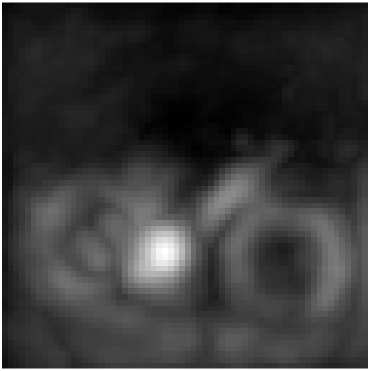}}\hspace{-0.3ex}
    \subfloat[\footnotesize{(c)Phasor field}] {\includegraphics[width=0.98in,height=30mm]{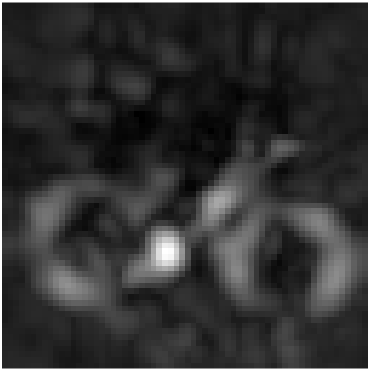}}\hspace{-0.3ex}
    \subfloat[\footnotesize{(d) F-K}]{\includegraphics[width=0.98in,height=30mm]{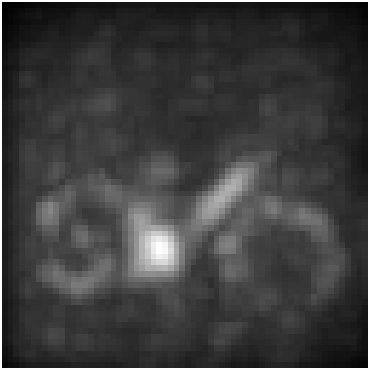}}\hspace{-0.3ex}
   \subfloat[\footnotesize{(e) SPIRAL}]{\includegraphics[width=0.98in,height=30mm]{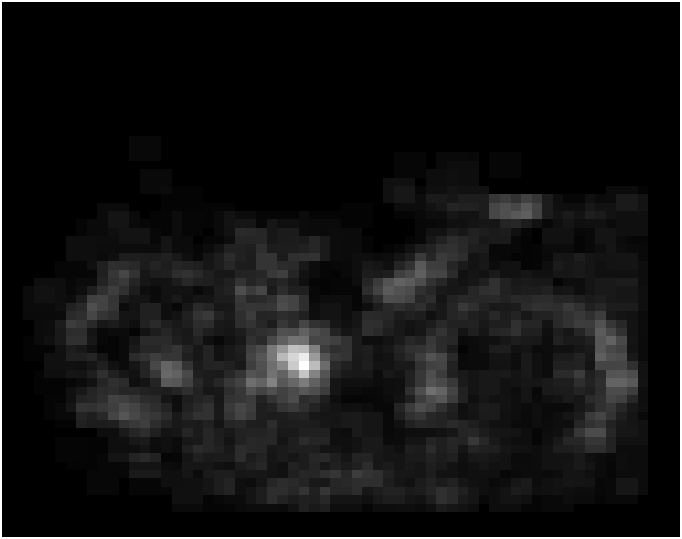}}\hspace{-0.3ex}
   \subfloat[\footnotesize{(f) Algorithm \ref{alg1}}]{\includegraphics[width=0.98in,height=30mm]{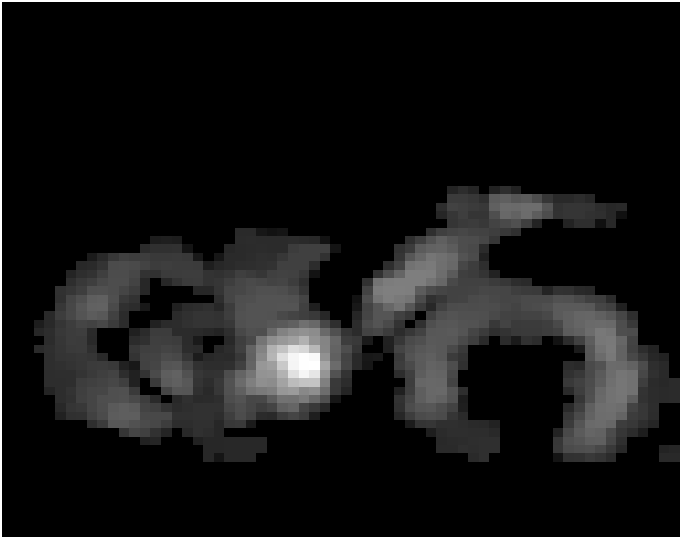}}\hspace{-0.3ex}
   \subfloat[\footnotesize{ (g) Algorithm \ref{alg2}}]{\includegraphics[width=0.98in,height=30mm]{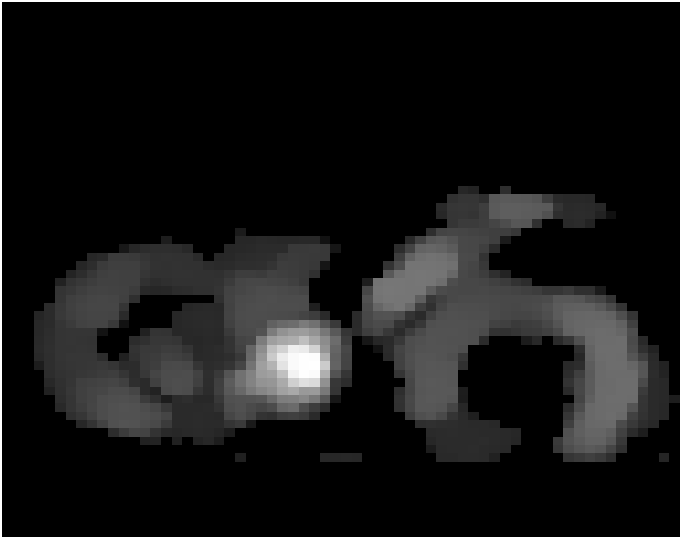}}\hspace{-0.3ex}\\
\captionsetup{font = small}	
 \caption{Comparison for reconstruction results of the bike (10 min), where the scanning points are of resolution $16\times 16$. The parameters are set as: $\mu$ = 1, $a = 2\times10^{-4}, b = 4\times10^{-5}$ for Algorithm \ref{alg1}; $\lambda = 35,a_{\tau} = 1\times10^{-4}, b_{\tau} = 1\times10^{-4},a_u = 6\times10^{-4}, b_u = 2\times10^{-4}$ for Algorithm \ref{alg2}.}
	\label{fig:bike}
\end{figure*}

In addition, we also apply our methods to another two scenes in the Stanford dataset, which are the outdoor and bike, respectively. The sizes of the raw measurement data are $128\times 128\times 2048$ and $512\times 512\times 2048$, respectively, while the wall size is $ 2\times 2$ m$^2$. The time resolution is cropped to 512 and each time bin spans $32$ ps. In the spatial dimension, the measurements are downsampled to 64 $\times$ 64. On this basis, we uniformly sample $16\times 16$ scanning points to reconstruct the two scenes. The comparison results are displayed in Fig. \ref{fig:outdoor} and Fig. \ref{fig:bike}, respectively. Similar to the previous experiments, the qualities of the images reconstructed by our methods are much better than the comparison methods. Although the difference between the images reconstructed by Algorithm \ref{alg1} and Algorithm \ref{alg2} is visually negligible for the outdoor scene, we can observe the advantages of dual-domain curvature regularization on bike scene, which is smoother than Algorithm \ref{alg1}.

\begin{figure*}
	\centering
        \captionsetup[subfloat]{labelsep=none,format=plain,labelformat=empty}
	\begin{minipage}[t]{0.18\linewidth}
		\centering
             \vspace{0.8in}
		\subfloat[(a)Ground truth]{\includegraphics[width=1.35in,height=42mm]{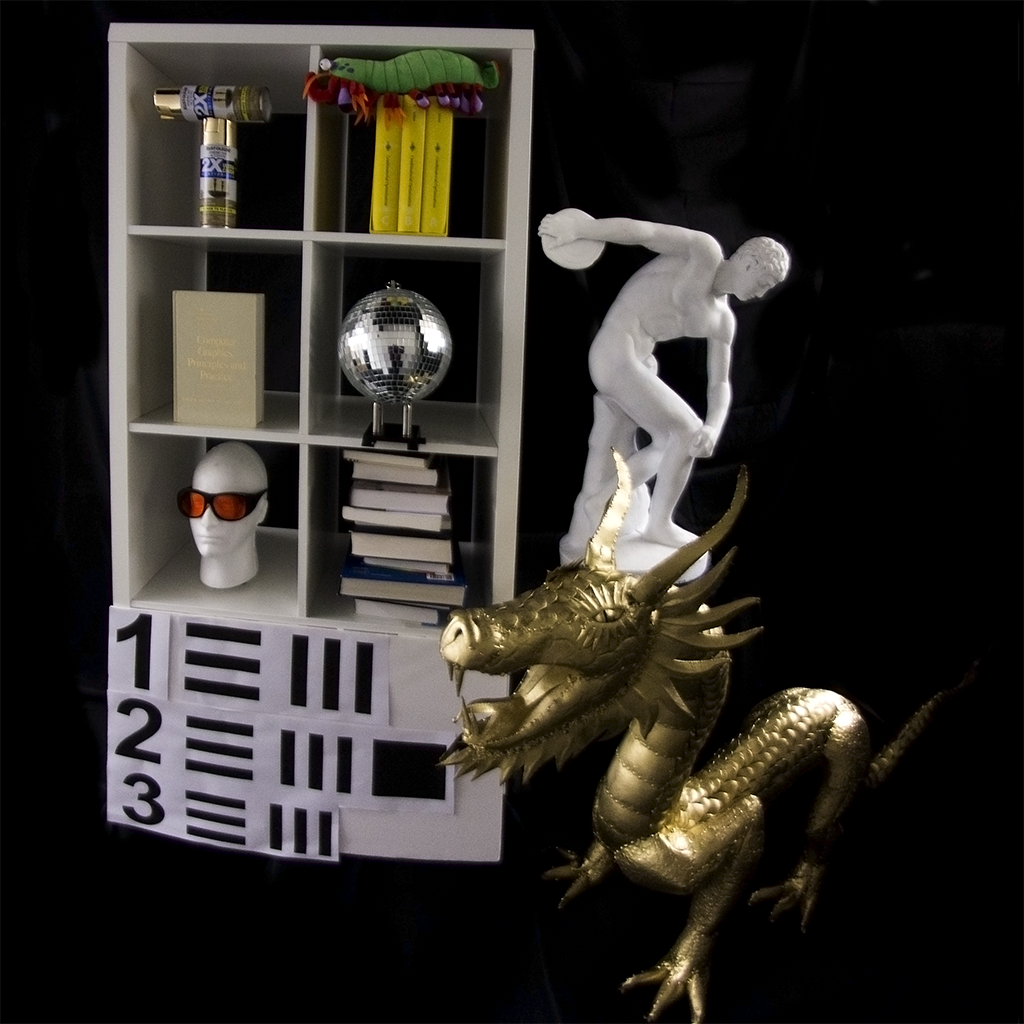}}\hspace{0.5ex}\\
	\end{minipage}%
	\begin{minipage}[t]{0.18\linewidth}
		\centering
             \subfloat{\includegraphics[width=1.5in,height=42mm]{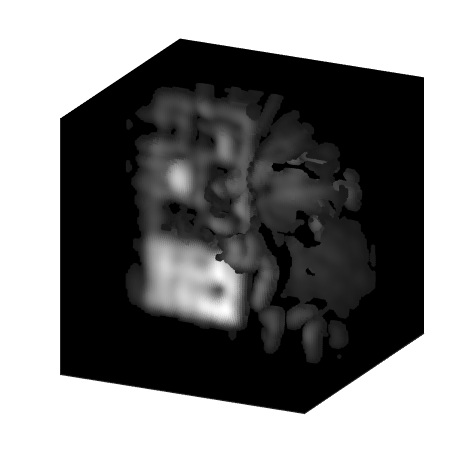}}\hspace{0.5ex}\\
             \vspace{-0.3in}
            \subfloat[(b)Phasor field]{\includegraphics[width=1.5in,height=42mm]{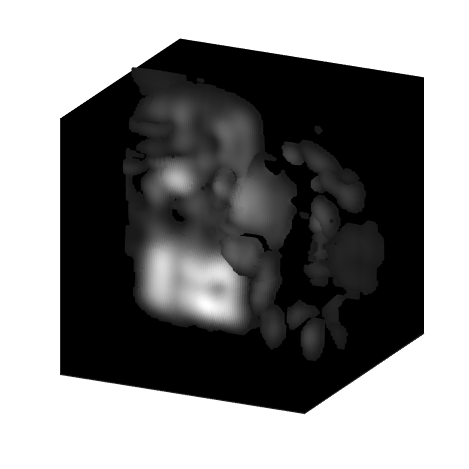}}\hspace{0.5ex}\\
		\vspace{0.02cm}
	\end{minipage}%
	\begin{minipage}[t]{0.18\linewidth}
		\centering
	      \subfloat{\includegraphics[width=1.5in,height=42mm]{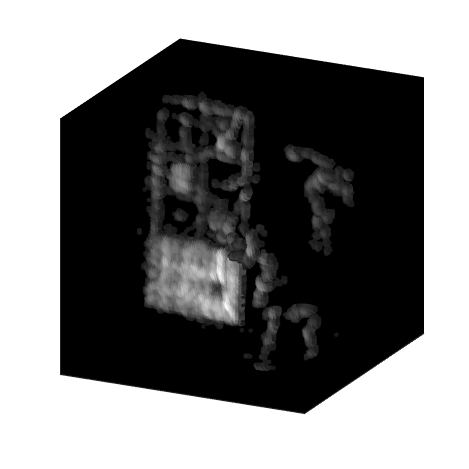}}\hspace{0.5ex}\\
        \vspace{-0.3in}
            \subfloat[(c)SPIRAL]{\includegraphics[width=1.5in,height=42mm]{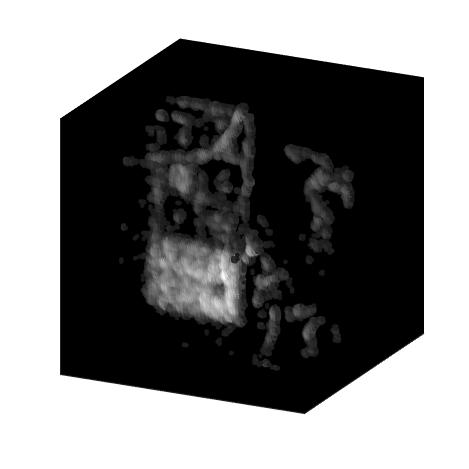}}\hspace{0.5ex}\\
	\end{minipage}%
	\begin{minipage}[t]{0.18\linewidth}
		\centering
	      \subfloat{\includegraphics[width=1.5in,height=42mm]{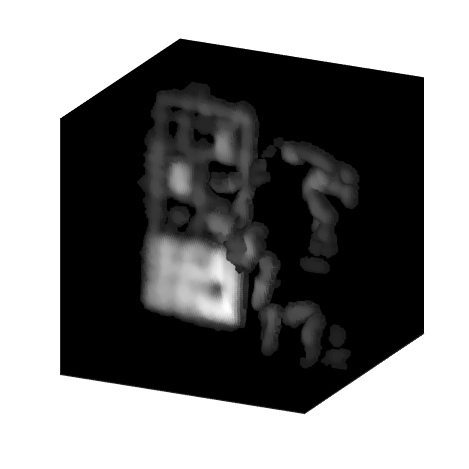}}\hspace{0.5ex}\\
       \vspace{-0.3in}
            \subfloat[(d)Algorithm \ref{alg1}]{\includegraphics[width=1.5in,height=42mm]{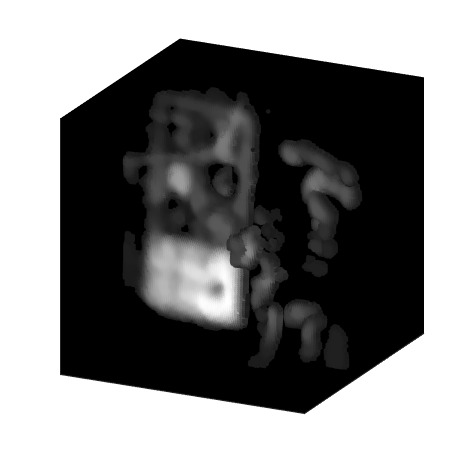}}\hspace{0.5ex}\\
	\end{minipage}%
        \begin{minipage}[t]{0.18\linewidth}
		\centering
	      \subfloat{\includegraphics[width=1.5in,height=42mm]{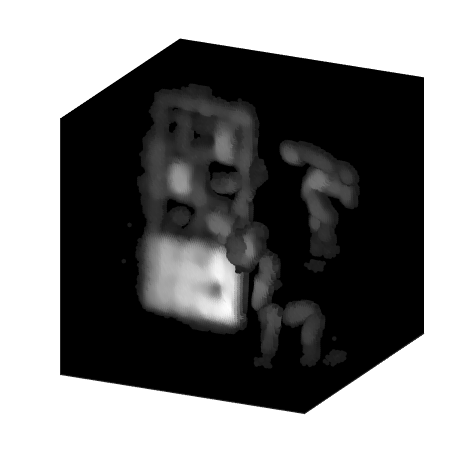}}\hspace{0.5ex}\\
       \vspace{-0.3in}
            \subfloat[(e)Algorithm \ref{alg2}]{\includegraphics[width=1.5in,height=42mm]{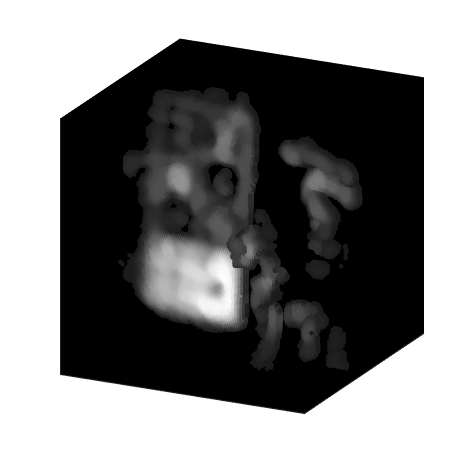}}\hspace{0.5ex}\\
	\end{minipage}%
	\centering
    \captionsetup{font = small}
	\caption{Comparison for reconstruction results of the teaser scene (180 min), where the scanning points are of resolution $24\times24$ (first row) and $16\times 16$ (second row). The 3D reconstruction results of Phasor field, SPIRAL, Algorithm \ref{alg1} and Algorithm \ref{alg2} methods are given here.}
        \label{fig:teaser}
\end{figure*}

Finally, we test different methods on a complex scene, i.e., teaser from the Stanford dataset. The raw measurement data of the teaser scenes is $512\times 512\times 2048$ and the wall size is $ 2\times 2$ m$^2$. The time resolution is cropped to 512 and each time bin spans $32$ps. Due to memory limitations, a total of $128\times 128$ scanning points are used by aggregating every $4\times4$ scanning points. Since data with an exposure time of 180min was used for testing, the scanning time of each point is about 0.66s. Fig. \ref{fig:teaser} presents the reconstruction results of both direct methods and iterative methods on the data obtained by scanning points of $24\times24$ and $16\times16$, respectively. As can be observed, both LCT and F-K could only reconstruct blurry contour information. Although the reconstruction results of the phasor field are better than the other two direct methods, they still lack detailed information about hidden objects and cannot discern specific structures. The effectiveness of iterative methods has been demonstrated when using a small number of scanning points. 
Specifically, we can see that the iterative method with $16\times16$ scanning points can reconstruct a better image than the phasor field with $24\times24$ scanning points.  We provide the exposure time and reconstruction time for different methods with the two combinations of scanning points in Table \ref{table:teaser time}. It can be observed that compared to the time spent on reconstruction, the cost of data acquisition is much higher. Although our method takes longer for reconstruction compared to the phasor field, we save a significant amount of exposure time when using $16\times16$ scanning points. To sum up, our method can spend less total time and provide better quality reconstruction results. 

\begin{table}[t]
\footnotesize
	\caption{Computational time for teaser scene with respect to $24\times24$ and $16\times16$ scanning points, where all time is recorded in second.}
	\label{table:teaser time}
	\centering
		\begin{tabular}{c|c|c|c|c}
			\Xhline{1.5pt}
{Scan Point}&{Exposure}&{Method}&{Reconstruct}&{Total time}\\
			\hline
\multirow{6}{*}{$24\times24$}&\multirow{6}{*}{380} 
            &LCT & 2.28 & 382.28 \\
			\cline{3-5}
			&&Phasor field& 3.02 & 383.02  \\
			\cline{3-5}
                &&F-K& 6.17 & 386.17 \\
			\cline{3-5}
			&&SPIRAL &541.44   &921.44   \\
            \cline{3-5}
            &&Algorithm \ref{alg1} & 21.37 &401.37   \\
            \cline{3-5}
            &&Algorithm \ref{alg2} & 49.50 &429.50  \\
			\cline{1-5}
			\multirow{6}{*}{$16\times16$} &\multirow{6}{*}{169} &LCT & 2.65 & 171.65 \\
			\cline{3-5}
			&&Phasor field& 2.91 & 171.91 \\
			\cline{3-5}
                &&F-K& 6.52 & 175.52 \\
			\cline{3-5}
			&&SPIRAL&  444.24 & 613.24 \\
            \cline{3-5}
            &&Algorithm \ref{alg1} & 21.34 & 190.34   \\
            \cline{3-5}
            &&Algorithm \ref{alg2} & 49.34 & 218.34 \\
			\Xhline{1.5pt}
		\end{tabular}
\end{table}

\begin{figure*}[t]
	\centering
        \captionsetup[subfloat]{labelsep=none,format=plain,labelformat=empty}
        \rotatebox{90}{\normalsize{\qquad\qquad \textbf{15s}}}
    \subfloat{\includegraphics[width=0.98in,height=30mm]{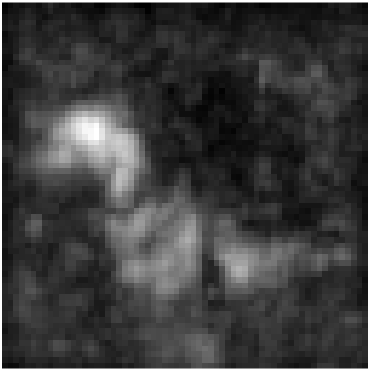}}\hspace{-0.3ex}
    \subfloat {\includegraphics[width=0.98in,height=30mm]{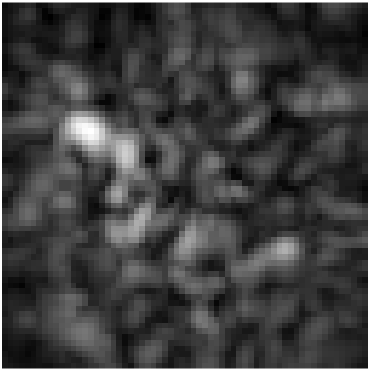}}\hspace{-0.3ex}
    \subfloat {\includegraphics[width=0.98in,height=30mm]{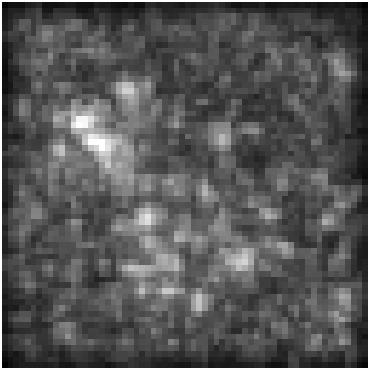}}\hspace{-0.3ex}
    \subfloat{\includegraphics[width=0.98in,height=30mm]{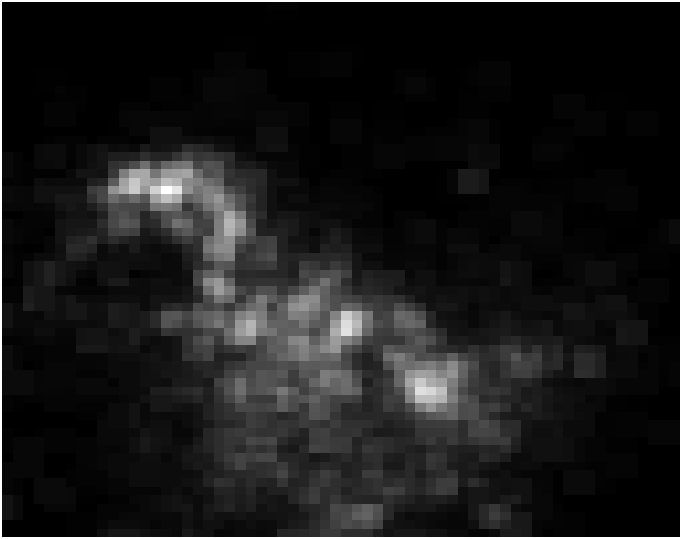}}\hspace{-0.3ex}
   \subfloat{\includegraphics[width=0.98in,height=30mm]{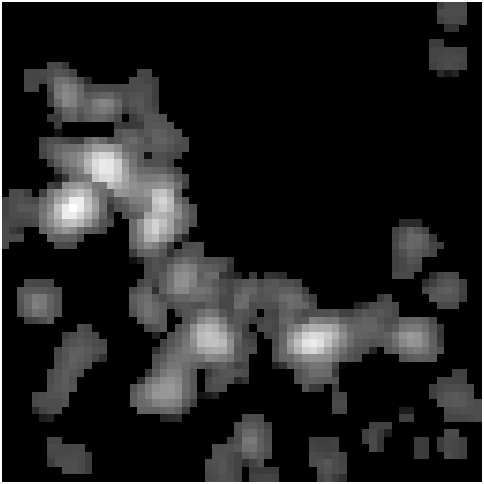}}\hspace{-0.3ex}
   \subfloat{\includegraphics[width=0.98in,height=30mm]{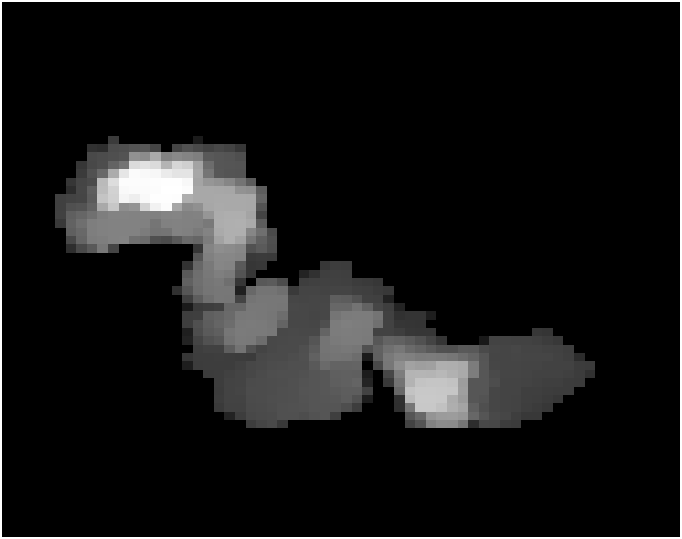}}\hspace{-0.3ex}
   \subfloat{\includegraphics[width=0.98in,height=30mm]{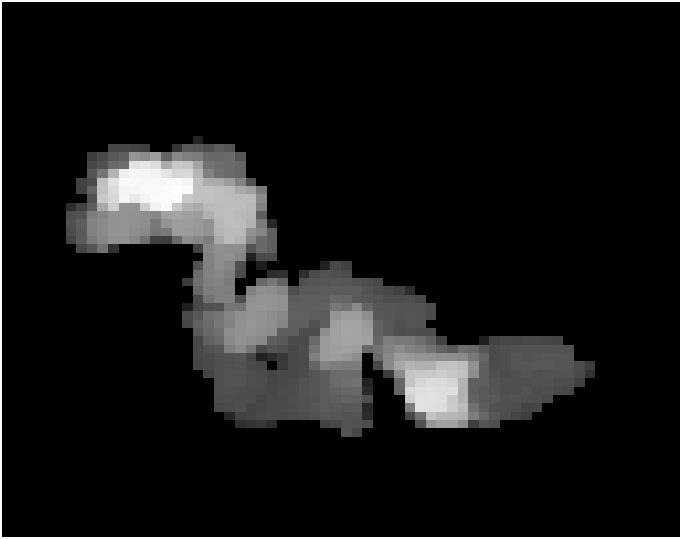}}\hspace{-0.3ex}\\   
    \vspace{0.2cm}
	\rotatebox{90}{\normalsize{\qquad\quad \textbf{60min}}}
    \subfloat[\footnotesize{(a) LCT}]{\includegraphics[width=0.98in,height=30mm]{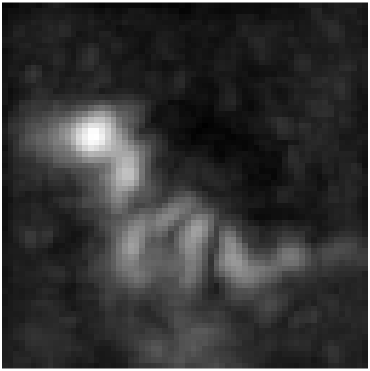}}\hspace{-0.3ex}
    \subfloat[\footnotesize{ (b) Phasor field}] {\includegraphics[width=0.98in,height=30mm]{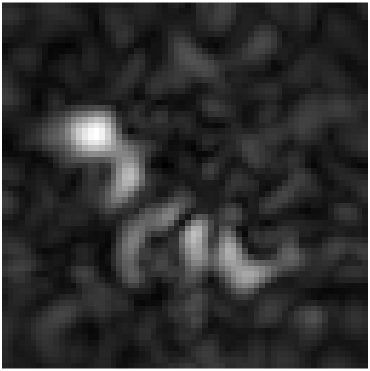}}\hspace{-0.3ex}
    \subfloat[\footnotesize{ (c) F-K}] {\includegraphics[width=0.98in,height=30mm]{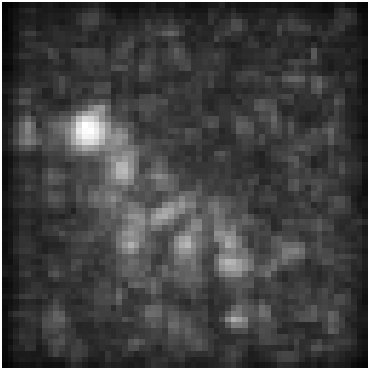}}\hspace{-0.3ex}
    \subfloat[\footnotesize{(d) SPIRAL}]{\includegraphics[width=0.98in,height=30mm]{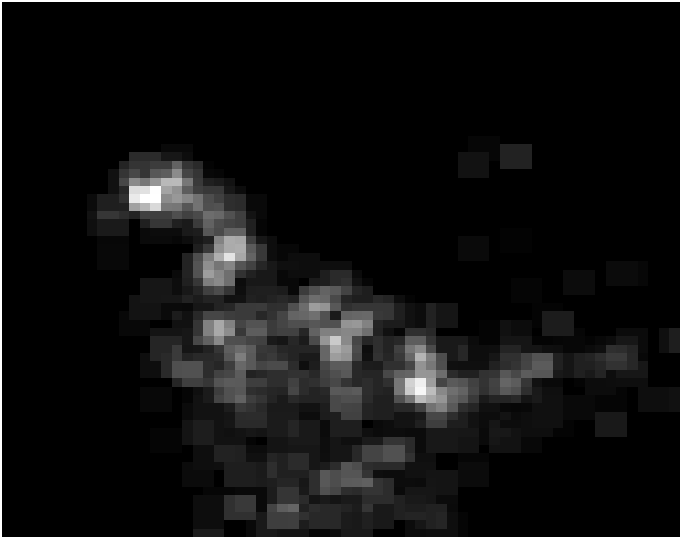}}\hspace{-0.3ex}
   \subfloat[\footnotesize{(e) SOCR}]{\includegraphics[width=0.98in,height=30mm]{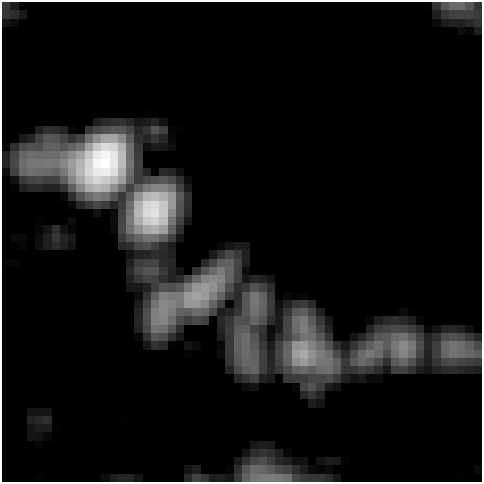}}\hspace{-0.3ex}
   \subfloat[\footnotesize{(f) Algorithm \ref{alg1}}]{\includegraphics[width=0.98in,height=30mm]{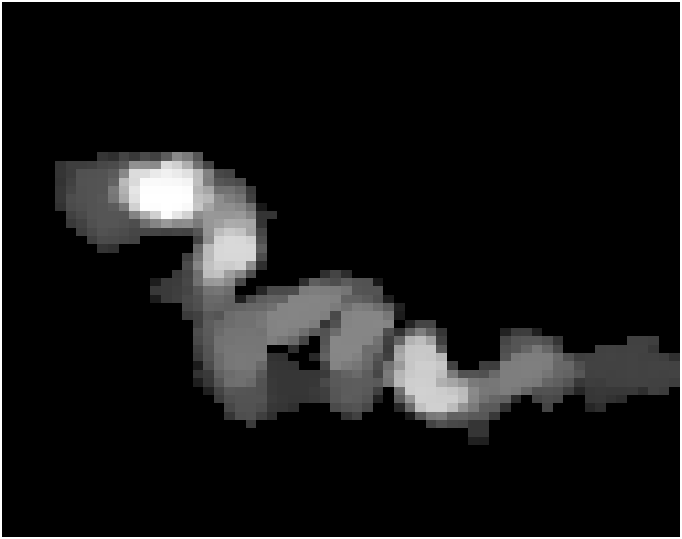}}\hspace{-0.3ex}
   \subfloat[\footnotesize{(g) Algorithm \ref{alg2}}]{\includegraphics[width=0.98in,height=30mm]{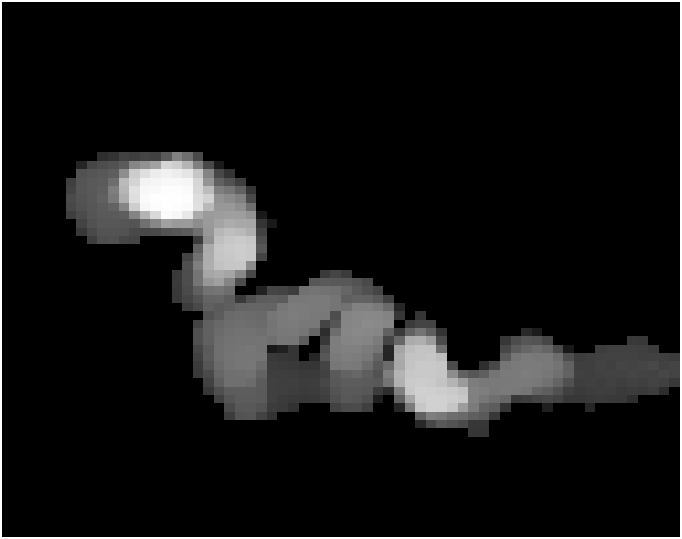}}\hspace{-0.3ex}
   \captionsetup{font = small}
	\caption{Comparison for reconstruction results of the dragon with an exposure time of 15s and 60min, where the scanning points are of resolution $16\times 16$. The parameters are set as: $\mu$ = 1, $a = 8\times10^{-4}, b = 5\times10^{-5}$ for data of exposure time 15s; $\mu$ = 1, $a = 1\times10^{-3}, b = 2.5\times10^{-4}$ for data of exposure time 60min.}
	\label{fig:dragon}
\end{figure*}

\begin{table*}[h]
	\centering
	\caption{The comparison of computational time among the comparison methods on the test data, where all time is recorded in second.}
        \label{table:time}
	\begin{tabular}{c|c|c|c|c|c|c|c|c}
		\Xhline{1.5pt}
		Image ID & Scanning points&LCT & Phasor field &F-K & SPIRAL & SOCR & Algorithm \ref{alg1} & Algorithm \ref{alg2} \\
		\hline
            Bowling &$64\times64$ & 0.28 & 0.35 & 0.84 &155& 876 & 5 &12 \\ \hline
		Bunny &$64\times64$ & 0.70 & 0.87 & 1.86 &156&3090 &6 &15 \\ \hline
		SU &$64\times64$ & 0.62 & 0.86 & 1.63 & 95&- &6 &15 \\ \hline
		Outdoor& $16\times16$ &0.59 & 0.73 & 1.49 &66 &2594 &6 &15 \\ \hline
		Bike & $16\times16$ & 0.57 & 0.76 & 1.59 &54 &- &6 &15 \\ \hline
		Dragon(15s) & $16\times16$ & 0.48 & 0.71 & 1.47 &71&3323&6 &14 \\ \hline
		Dragon(60min) & $16\times16$ & 0.47 & 0.69 & 1.56 &53 &2372 &6 &14 \\
		\Xhline{1.5pt}
	\end{tabular}
	\label{time}
\end{table*}

\subsection{Experiments on data with different exposure time}
In this subsection, we verify the performance of our curvNLOS method on the dragon scene of spatial resolution 64$\times$64, where two measurements were captured with different exposure times, i.e., $15$s and $60$min respectively. The shorter the total exposure time is, the fewer photons are captured at each point, and the data is more affected by noises. The reconstruction results are shown in Fig. \ref{fig:dragon}. It can be seen that except for our curvNLOS, all the comparison methods cannot reconstruct satisfactory images for the data estimated by short exposure time. The images reconstructed by Phasor field and F-K methods are blurry and unrecognizable. The images reconstructed by LCT and SPIRAL methods can be roughly identified, but contain a large amount of noise. The reconstruction of the SOCR is also greatly affected by noise and structural loss. Although two legs of the dragon are connected, our object-domain Algorithm \ref{alg1} still produces reconstruction results with much better quality. On the other hand, the reconstruction results of all comparison methods are significantly improved on the long exposure time data. We can observe the rough shape of the dragon in the reconstructed images of LCT, Phasor field, F-K, SPIRAL, and SOCR, but they are still greatly affected by noises. The result of our method is the best. By increasing the exposure time, the reconstruction quality is improved, especially the edge information.

\subsection{Computational time comparison}
In this subsection, we compare the computational time among the comparison methods on different scenes, which are exhibited in Table \ref{time}. We can see that the iterative reconstruction methods, i.e.,  SPIRAL, SOCR, and our curvNLOS, consume more time than the direct reconstruction methods, i.e., LCT, Phasor field and F-K. As can be observed, the computational time of SPIRAL varies with data dimensions and scenes. It converges fast in the bike scene, while it consumes much more time to reconstruct the Bunny scene. 
Furthermore, it reveals that the computational cost of SOCR is much higher due to the signal-object collaborative regularization. Although we implemented the parallel codes with 12 workers, the reconstruction time is still very long. For our curvNLOS, due to the GPU computation, it consumes the least computational time among the three iterative methods. Since there is no inner iteration in our approaches, the computational time is relatively stable without varying too much for measured data of the same dimensions. However, the computational costs of our Algorithm \ref{alg1} and Algorithm \ref{alg2} are still much higher than direct methods, which are 7 times and 18 times more than the Phasor field. We may further improve the efficiency by using better GPU cards in the future to mimic the gaps between our methods and direct reconstruction methods.


\subsection{Evaluation on non-confocal data}

In confocal laser imaging, the laser source and detector share the same focal point, which can maximize the detection of directly reflected light from the target's surface. However, it may suppress the light scattered by hidden objects, making it difficult to extract useful information. Non-confocal imaging decouples the focus of the light source and detector, which can effectively collect light undergoing multiple scattering interactions. Although the signal is weak, it contains valuable information about structures hidden beneath the surface or occlusion on the beam path. By capturing greater scattering diversity, non-confocal methods can identify missing information from confocal systems in an extended field of view. In the following, we validate the advantages of our curvNLOS method on the non-confocal data, where the letter `K' from NLOS Benchmark dataset with a visible wall of size $0.512\times0.512 m^2$ and a photon propagation distance of $0.001$ meters per second. In our experiments, we employ the `K' data downsampled to $64\times 64$ as provided by \cite{liu2021non}, which contains 1024 time bins. 
On the one hand, we compared our method with SOCR's non-confocal solver. On the other hand, we converted the non-confocal data into confocal data by the midpoint approximation method \cite{lindell2019wave} and compared our method with LCT, phasor field, F-K, and SPIRAL. 

\begin{figure}[t]
	\centering
 \captionsetup[subfloat]{labelsep=none,format=plain,labelformat=empty}
        \rotatebox{90}{\scriptsize{\qquad~\textbf{64$\times$64}}}
    \subfloat{\includegraphics[width=0.56in,height=18mm]{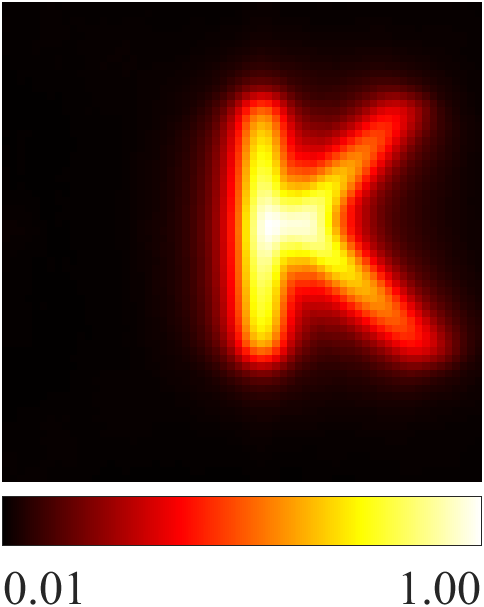}}
    \subfloat{\includegraphics[width=0.56in,height=18mm]{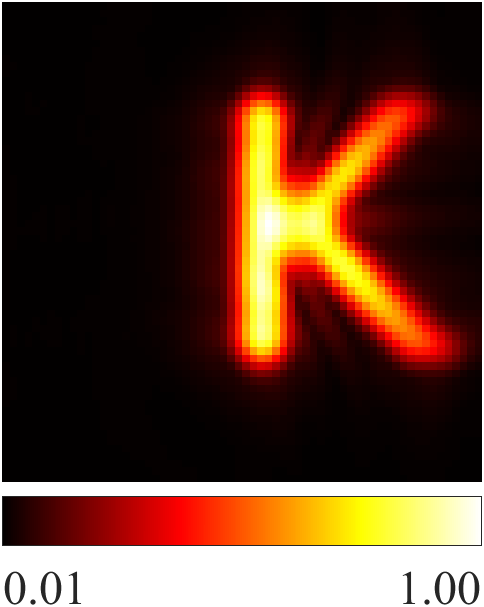}}
    \subfloat{\includegraphics[width=0.56in,height=18mm]{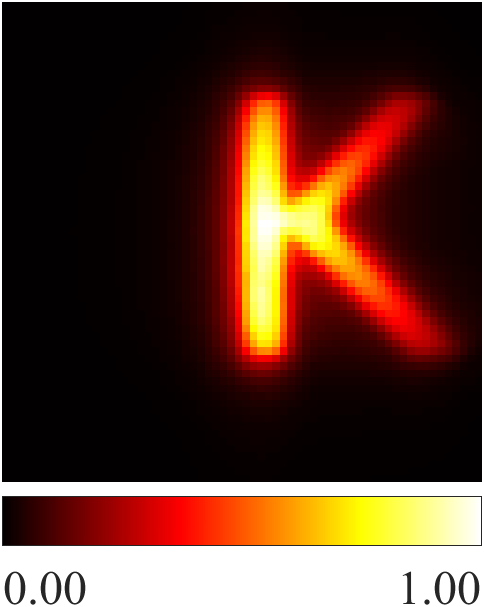}}
    \subfloat{\includegraphics[width=0.56in,height=18mm]{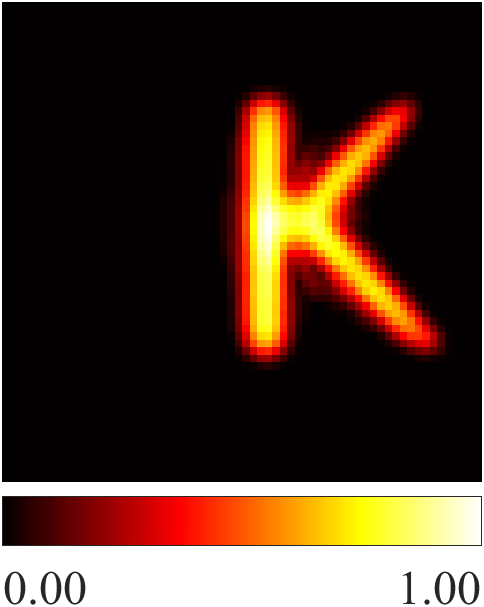}}
    \subfloat{\includegraphics[width=0.56in,height=18mm]{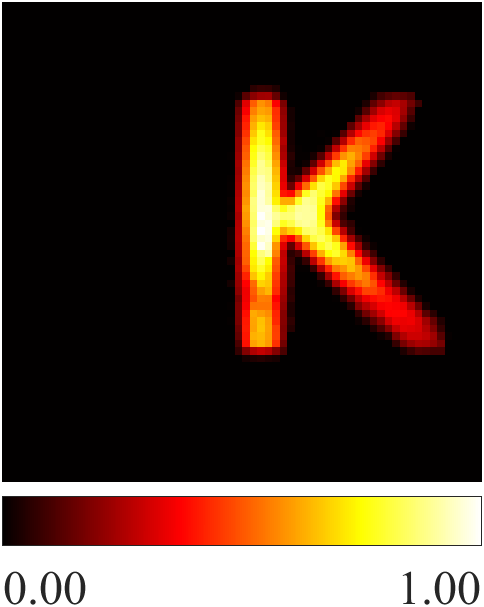}}
    \subfloat{\includegraphics[width=0.56in,height=18mm]{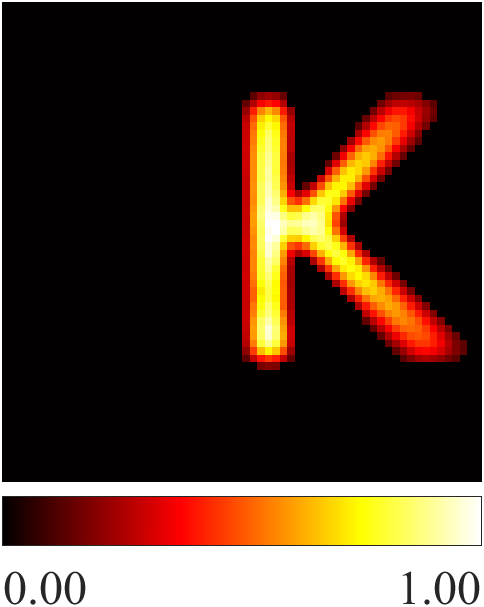}}\\

    \rotatebox{90}{\scriptsize{\qquad~\textbf{16$\times$16}}}
    \subfloat{\includegraphics[width=0.56in,height=18mm]{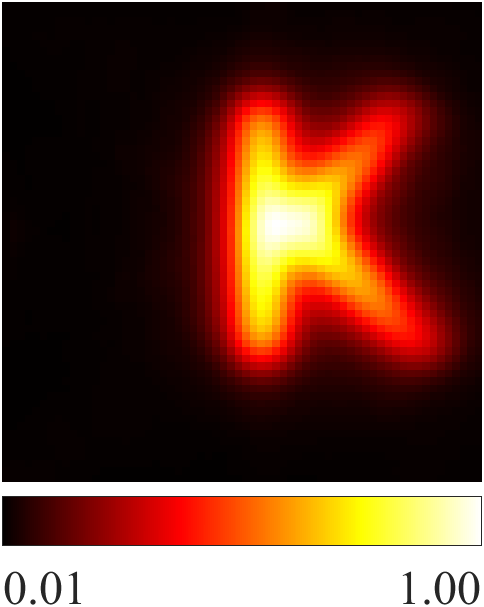}}
    \subfloat{\includegraphics[width=0.56in,height=18mm]{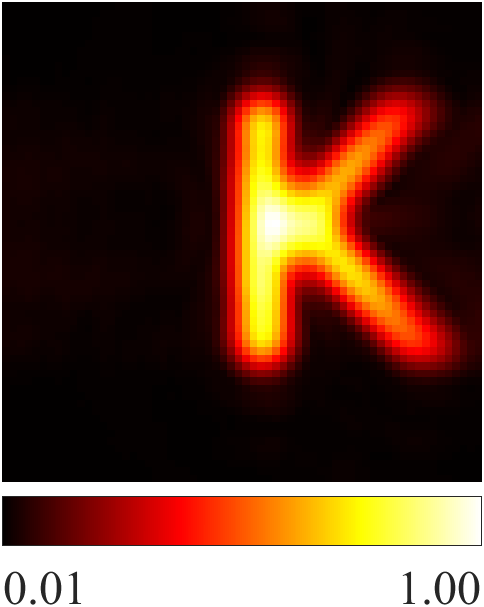}}
    \subfloat{\includegraphics[width=0.56in,height=18mm]{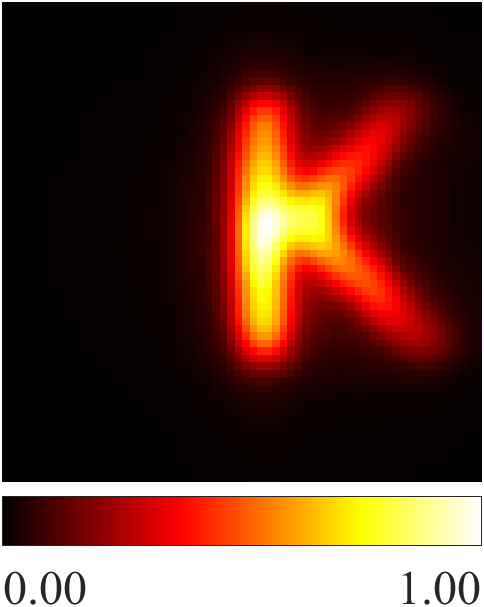}}
    \subfloat{\includegraphics[width=0.56in,height=18mm]{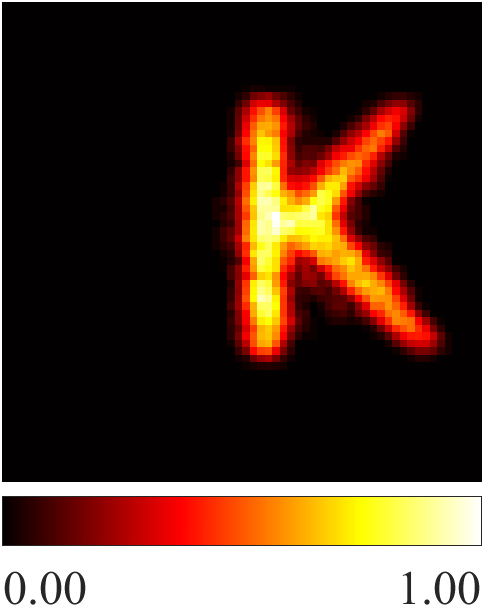}}
    \subfloat{\includegraphics[width=0.56in,height=18mm]{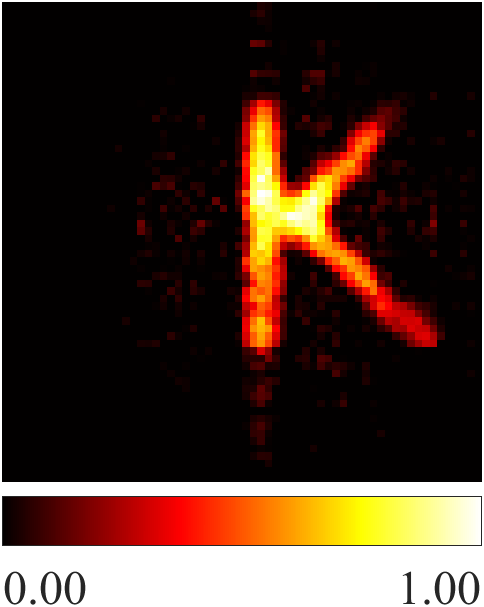}}
    \subfloat{\includegraphics[width=0.56in,height=18mm]{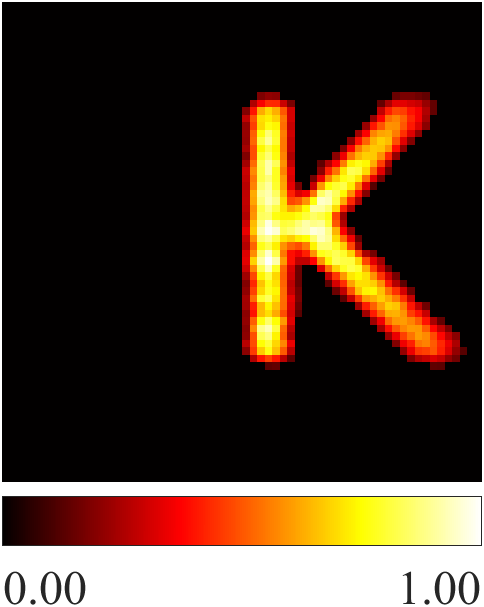}}\\

    \rotatebox{90}{\scriptsize{\qquad\quad\textbf{8$\times$8}}}
    \subfloat{\includegraphics[width=0.56in,height=18mm]{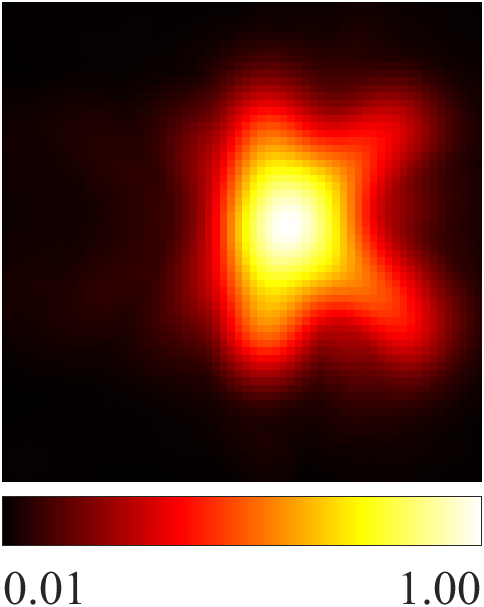}}
    \subfloat{\includegraphics[width=0.56in,height=18mm]{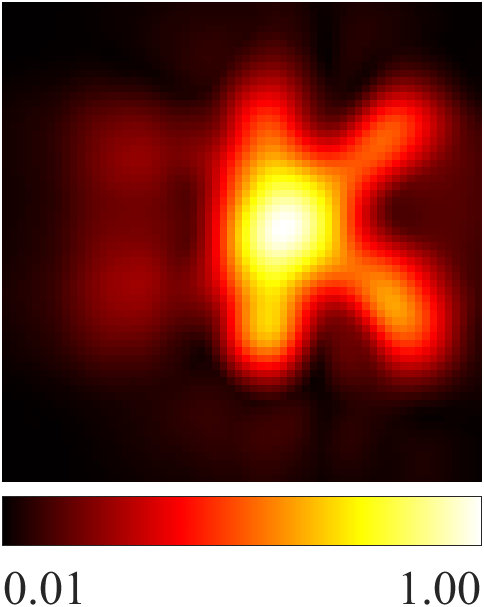}}
    \subfloat{\includegraphics[width=0.56in,height=18mm]{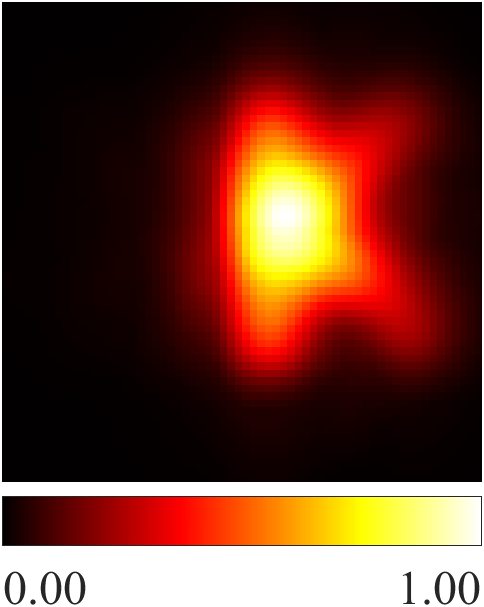}}
    \subfloat{\includegraphics[width=0.56in,height=18mm]{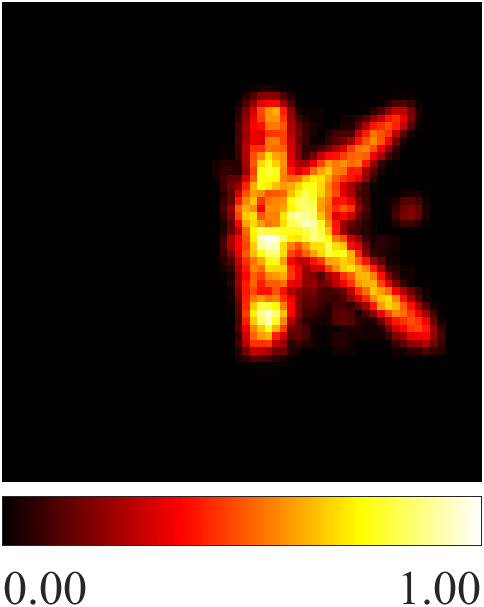}}
    \subfloat{\includegraphics[width=0.56in,height=18mm]{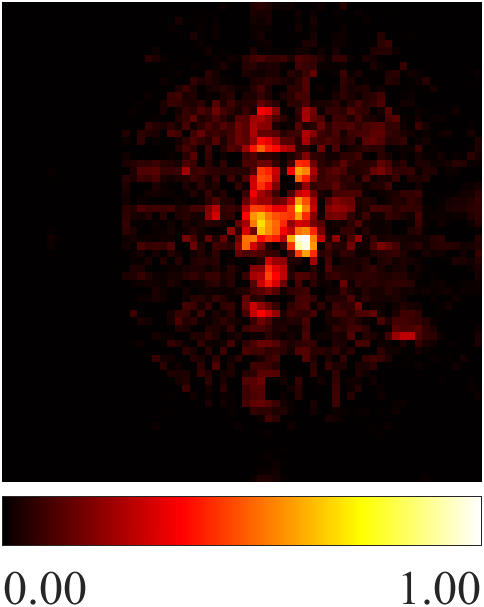}}
    \subfloat{\includegraphics[width=0.56in,height=18mm]{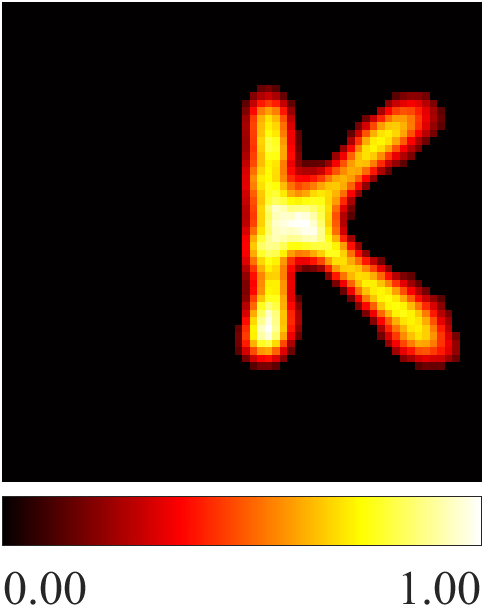}}\\

    \rotatebox{90}{\scriptsize{\qquad\quad\textbf{4$\times$4}}}
    \subfloat[\tiny{\textbf{(a) LCT}}]{\includegraphics[width=0.56in,height=18mm]{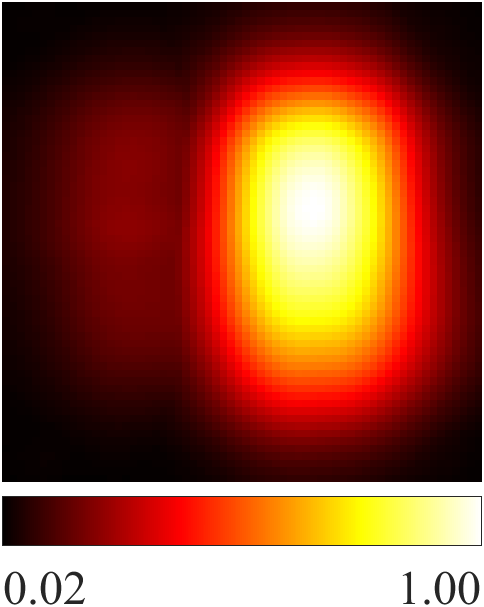}}
    \subfloat[\tiny{\textbf{(b) Phasor field}}]{\includegraphics[width=0.56in,height=18mm]{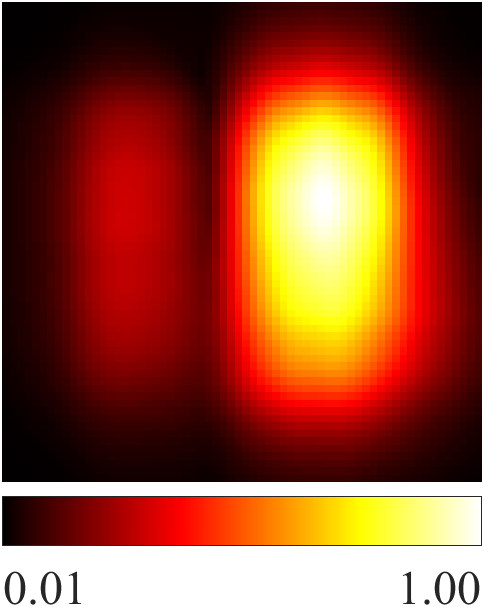}}
    \subfloat[\tiny{\textbf{(c) F-K}}]{\includegraphics[width=0.56in,height=18mm]{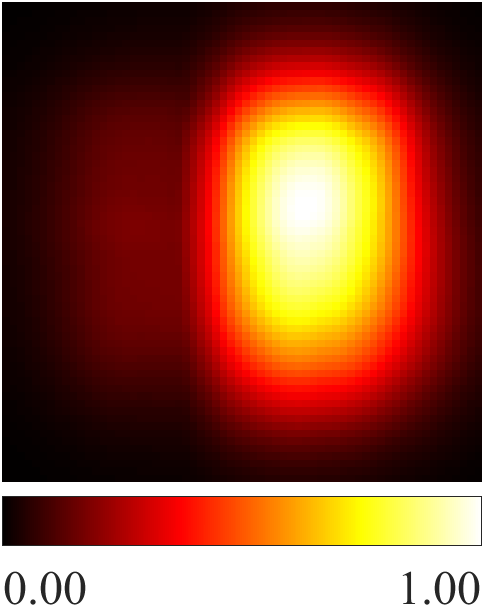}}
    \subfloat[\tiny{\textbf{(d) SPIRAL}}]{\includegraphics[width=0.56in,height=18mm]{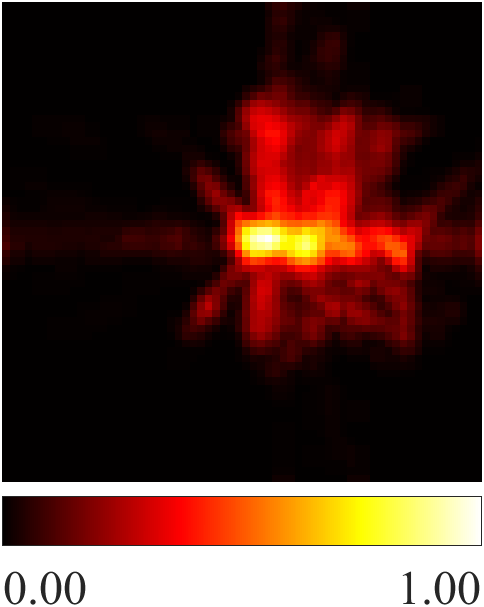}}
    \subfloat[\tiny{\textbf{(e) SOCR}}]{\includegraphics[width=0.56in,height=18mm]{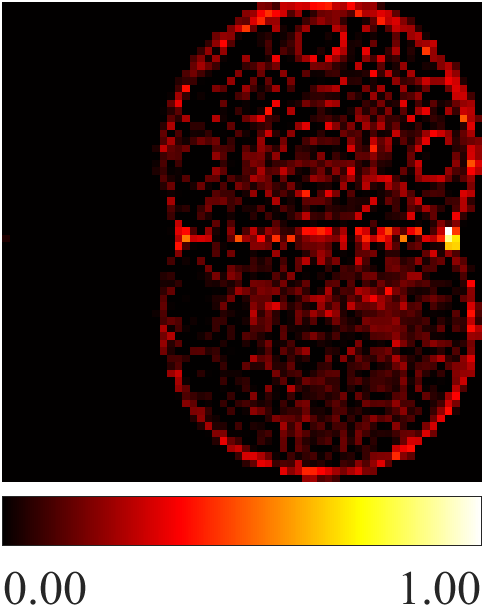}}
    \subfloat[\tiny{\textbf{(f) CurvNLOS}}]{\includegraphics[width=0.56in,height=18mm]{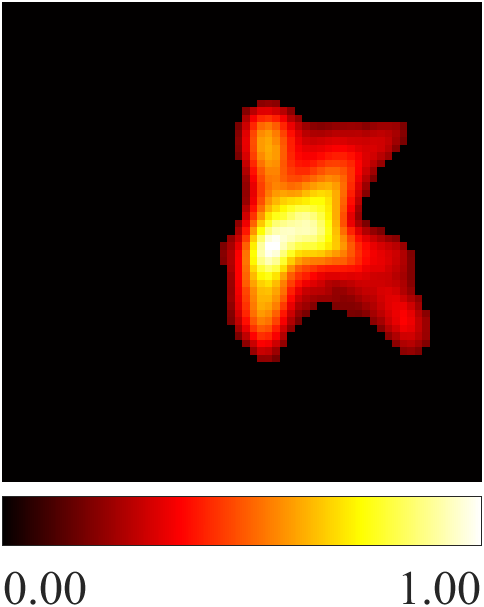}}
    \captionsetup{font = small}
    \caption{Comparison for reconstruction results of the K scene (non-confocal) using the scanning points of $64\times 64$, $16\times 16$, $8\times8$, $6\times 6$, and $4\times4$ from left to right. The parameters are set as: $a=6\times10^{-6}, b=8\times 10^{-6}$ ($64\times 64$),  $a=4\times10^{-4}, b=8\times 10^{-4}$ ($16\times 16$),  $a=6\times10^{-4}, b=4\times 10^{-4}$ ($8\times 8$), $a=6\times10^{-2}, b=8\times 10^{-2}$ ($4\times 4$). }
	\label{fig:NC}
\end{figure}

Different scanning points are chosen for evaluation, the reconstruction results of which are provided in 
Fig. \ref{fig:NC}. As can be observed, both our method and SOCR can reconstruct higher-quality images under full sampling, demonstrating the advantages of non-confocal data compared to LCT, Phasor field, F-K, and SPIRAL. For the scanning point of $16\times16$, the reconstruction results of SOCR showed certain degradation, and the brightness of the reconstructed object significantly decreased. The reconstruction results of LCT, Phasor field, F-K, and SPIRAL become blurred and lack details. Our method gives the best reconstruction quality with relatively clear contours. When the scanning point drops to $4\times4$, other methods are no longer able to reconstruct the meaningful image, and our method can still distinguish the shape of K.

\section{Conclusion and Future Works}
\label{sec6}
In this paper, we introduced the curvature regularization models for both confocal and non-confocal NLOS reconstruction problems with sparse scanning points. The sparse scanning can effectively shorten the acquisition time, but it also leads to the failure of reconstruction methods. The curvature regularization used in the object domain and original signal domain can not only restore the smooth surface of the hidden objects but also the continuous signals. Fast numerical algorithms were proposed for solving the high-order curvature minimization problems, where the curvature function was regarded as the adaptive weight for the total variation to ease the computation, and the linearization technique was used to accelerate the convergence.
Extensive numerical experiments were conducted on both synthetic and real data. Compared to state-of-the-art direct and iterative NLOS reconstruction methods, our curvNLOS was shown with better reconstruction qualities for different scenes, demonstrating the effectiveness of the curvature in recovering the three-dimensional surfaces. The results showed our curvNLOS can reconstruct the 3D hidden scenes with $64\times64$ spatial resolution by the measurements of $4\times4$ sampling points. Besides that, thanks to the GPU implementation, our curvNLOS performed much faster than other iterative reconstruction methods, which can facilitate its use in real applications.

Although this work improves both reconstruction quality and computational efficiency, NLOS imaging reconstruction is still a very challenging problem. The existing reconstruction methods suffer from poor spatial resolution, noise sensitivity, and poor real-time realization. Thus, our future work includes using the super-resolution method and deep learning technique to obtain more efficient NLOS reconstruction methods. Furthermore, although uniform sampling can be used to reduce scan points, it lacks integration with the physical imaging process and needs to be optimized in the future.


\ifCLASSOPTIONcompsoc
  


\ifCLASSOPTIONcaptionsoff
  \newpage
\fi



\bibliographystyle{IEEEtran}
\bibliography{nlos}
%

%





\vspace{-0.5in}
\begin{IEEEbiography}
[{\includegraphics[width=1in,height=1.25in,clip,keepaspectratio]{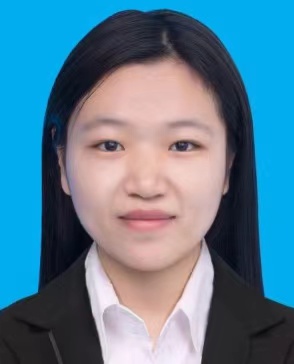}}]
Rui Ding received the B.S. degree in Mathematics from Harbin Engineering University (HEU) in 2019. She received master's degree from the Center for Applied Mathematics of Tianjin University in 2023. Her research interests include non-line-of-sight imaging and image reconstruction.
\end{IEEEbiography}
\vspace{-0.5in}
\begin{IEEEbiography}
[{\includegraphics[width=1in,height=1.15in,clip,keepaspectratio]{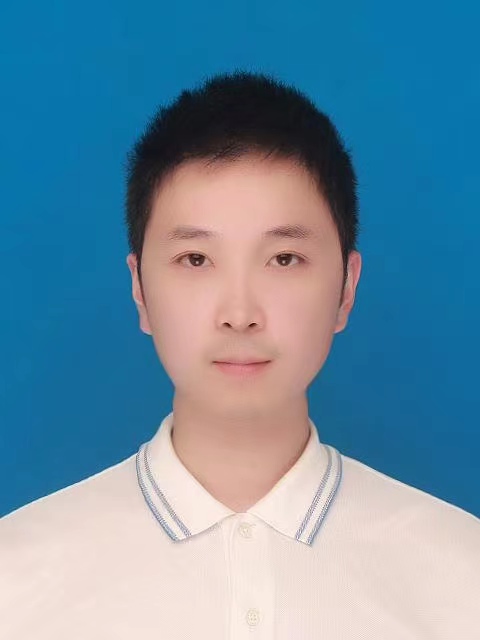}}]
Juntian Ye received a B.S. degree in physics from the University of Science and Technology of China (USTC), Hefei, Anhui, China, in 2018. He is currently working toward a Ph.D. degree with the Department of Modern Physics, USTC. His research interests include non-line-of-sight imaging, single-photon LiDAR, and other computational imaging systems.
\end{IEEEbiography}
\vspace{-0.5in}
\begin{IEEEbiography}
[{\includegraphics[width=1in,height=1.25in,clip,keepaspectratio]{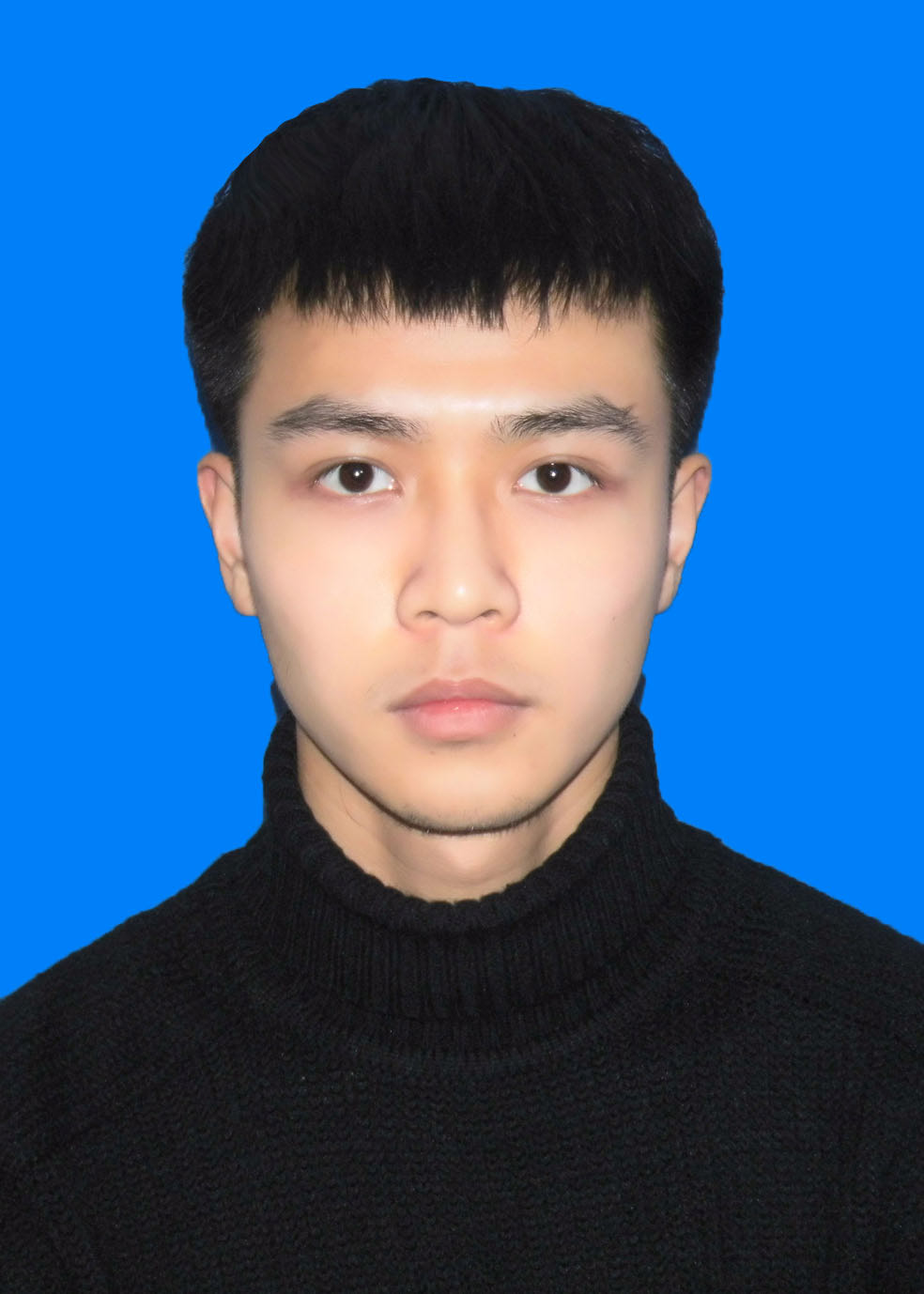}}]
Qifeng Gao is currently pursuing a Ph.D in Center for Applied Mathematics of Tianjin University. His work focuses on deep learning based methods applied in the imaging fields of computed tomography imaging, fluorescence microscopy imaging, and non-line-of-sight imaging.
\end{IEEEbiography}
\vspace{-0.5in}
\begin{IEEEbiography}
[{\includegraphics[width=1in,height=1.25in,clip,keepaspectratio]{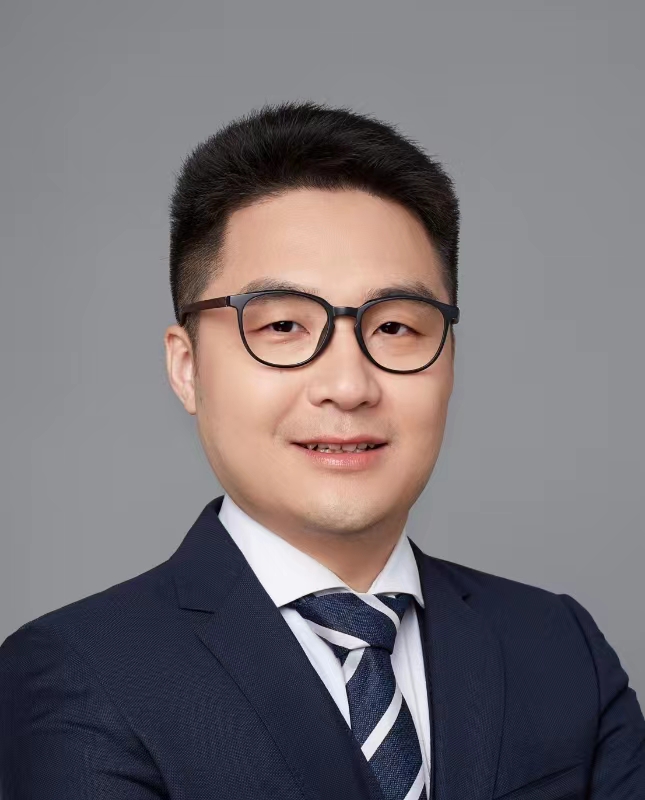}}]
Feihu Xu has been a Full Professor of Physics at USTC since 2021. Before joining USTC in 2017, he was a Postdoctoral Associate at MIT in 2015-2017. He received an M.A.Sc and Ph.D. from the University of Toronto in 2011 and 2015. He works on quantum information science and single-photon imaging and has co-authored more than 100 journal papers including RMP, Nature, Nat Photon, PRL/X, etc. He is a fellow of Optica and IOP. He is the recipient of the International Quantum Technology Early Career Scientist Award, Changjiang Scholar, Xplorer Prize, MIT Technology Review 35 Innovators Under 35 of China, OCPA Outstanding Dissertation Award, and QCrypt Best Paper Award.
\end{IEEEbiography}
\vspace{-0.5in}
\begin{IEEEbiography}
[{\includegraphics[width=1in,height=1.25in,clip,keepaspectratio]{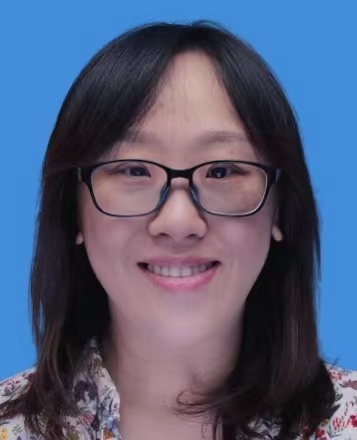}}]
Yuping Duan is a full professor at the School of Mathematical Sciences of Beijing Normal University (BNU). Before joining BNU, she was a professor at Tianjin University in 2015 to 2023, and a research scientist at I2R, A*STAR in 2012 to 2015. She received her Ph.D. from Nanyang Technological University in 2012. Her research interests are image processing and computer vision, variational methods, and deep learning methods.
\end{IEEEbiography}
\end{document}